\documentclass[conference]{IEEEtran}
\IEEEoverridecommandlockouts

\usepackage{cite}
\usepackage{booktabs}
\usepackage{comment}
\usepackage{multirow}
\usepackage{hyperref}
\usepackage{orcidlink}
\usepackage{amsmath,amssymb,amsfonts}
\usepackage{algorithmic}
\usepackage{graphicx}
\usepackage{textcomp}
\usepackage{xcolor}
\usepackage{tikz}
\usepackage{hyperref }
\usepackage{siunitx}
\usepackage{todonotes}
\usepackage{fontawesome}

\usepackage[final,commandnameprefix=ifneeded]{changes}
\definechangesauthor[name={Tim Schwabe}, color=red]{TS}
\definechangesauthor[name={John Maek}, color=orange]{JM}

\usetikzlibrary{
    arrows.meta,
    positioning,
    shapes.geometric,
    shapes.misc,
    calc,
    backgrounds,
    fit,
    decorations.pathreplacing,
    decorations.markings
}
\usepackage{standalone}
\usepackage{bm}
\definecolor{thetadark}{HTML}{64748B}      
\definecolor{primblue}{HTML}{3B82F6}       
\definecolor{modelblue}{HTML}{6366F1}      
\definecolor{globalgreen}{HTML}{059669}    
\definecolor{scoregold}{HTML}{D97706}      
\definecolor{consteal}{HTML}{0D9488}       
\definecolor{uprarrow}{HTML}{DC2626}       
\definecolor{xidash}{HTML}{8B5CF6}         
 
\definecolor{panelbg}{HTML}{F8FAFC}        
\definecolor{arrowlabel}{HTML}{475569}     
\definecolor{titledark}{HTML}{1E293B}      
\definecolor{levelbox}{HTML}{94A3B8}       
 
\definecolor{gnnbox}{HTML}{0EA5E9}         
\definecolor{outputcol}{HTML}{F97316}      
\definecolor{pretrain}{HTML}{DB2777}       
 \usepackage{xcolor}
\definecolor{poolWhite}{HTML}{FFFFFF}
\definecolor{uprGray}{HTML}{EEEEEE}
\definecolor{dimBlue}{HTML}{D4E6F1}
\definecolor{globOrg}{HTML}{FDE8D1}
\definecolor{phiBlue}{HTML}{2166AC}
\definecolor{concatPurp}{HTML}{E8DAEF}
\definecolor{rhoPlum}{HTML}{5E3C99}
\definecolor{outRed}{HTML}{E74C3C}
\definecolor{arrowGray}{HTML}{444444}
\definecolor{labelGray}{HTML}{222222}
\definecolor{umarrow}{HTML}{DC2626}
\definecolor{bracketCol}{HTML}{2166AC}
\definecolor{uprGray}{HTML}{EEEEEE}
\definecolor{umarrow}{HTML}{DC2626}
\definecolor{dimBlue}{HTML}{3B82F6}
\definecolor{globOrg}{HTML}{FDE8D1}
\definecolor{attnGreen}{HTML}{059669}
\definecolor{logPlum}{HTML}{5E3C99}
\definecolor{concatPurp}{HTML}{E8DAEF}
\definecolor{consteal}{HTML}{0D9488}
\definecolor{arrowGray}{HTML}{444444}
\definecolor{bracketCol}{HTML}{2166AC}
\newtheorem{definition}{Definition}
\def\BibTeX{{\rm B\kern-.05em{\sc i\kern-.025em b}\kern-.08em
    T\kern-.1667em\lower.7ex\hbox{E}\kern-.125emX}}

\newif\ifarxiv
\arxivtrue       

\newcommand{\arxivonly}[1]{\ifarxiv#1\fi}

\begin{document}

\title{Estimating Inconsistency Response Surfaces under Uncertainty in Cyber-Physical System Development\\
\arxivonly{
\thanks{Extended version of the paper accepted at the IEEE International Conference on Data Mining (ICDM) 2026; additional material is provided in the appendices. \copyright~2026 IEEE. 
Personal use of this material is permitted. 
Permission from IEEE must be obtained for all other uses, in any current or future media, including reprinting/republishing this material for advertising or promotional purposes, creating new collective works, for resale or redistribution to servers or lists, or reuse of any copyrighted component of this work in other works.}}
\thanks{Funded by the Deutsche Forschungsgemeinschaft (DFG, German Research Foundation) – SFB 1608 – 501798263.}
}

\author{
\IEEEauthorblockN{Johannes Mäkelburg\, \orcidlink{0009-0001-3821-7817}}
\IEEEauthorblockA{
\textit{Technische Universität München} \\
Munich, Germany \\
\href{mailto:johannes.maekelburg@tum.de}{johannes.maekelburg@tum.de}
}
\and
\IEEEauthorblockN{Tim Schwabe\, \orcidlink{0009-0009-7957-603X}}
\IEEEauthorblockA{
\textit{Technische Universität München} \\
Munich, Germany \\
\href{mailto:tim.schwabe@tum.de}{tim.schwabe@tum.de}
}
\and
\IEEEauthorblockN{Maribel Acosta\,\orcidlink{0000-0002-1209-2868}}
\IEEEauthorblockA{
\textit{Technische Universität München} \\
Munich, Germany \\
\href{mailto:maribel.acosta@tum.de}{maribel.acosta@tum.de}
}
}

\maketitle

\begin{abstract}
Cyber-Physical Systems (CPS) are commonly represented through multiple interconnected models. 
During development, CPS consistency requires that shared model elements remain compatible across these models. 
Uncertainty, for example, due to sensor noise or model abstraction, changes the admissible values of model elements and can introduce inconsistencies, i.e., situations in which models can no longer be jointly satisfied. 
While existing approaches can determine consistency for a given uncertainty configuration, they provide limited support for systematically exploring, analyzing, and explaining inconsistency across large uncertainty spaces.
We address this challenge by reformulating inconsistency as an intervention response modeling problem. 
Using Saltelli sampling and multi-fidelity Monte Carlo estimation, we generate intervention-response datasets and train a surrogate model that directly predicts inconsistency from the propagated uncertainty geometry.
Experiments on 48 scenarios and 10 CPS domains show that the surrogate matches Monte Carlo estimates while reducing evaluation time from milliseconds to microseconds, enabling orders-of-magnitude more response-surface evaluations within fixed computational budgets.
Building on the learned response surfaces, we perform sensitivity analysis to identify dominant uncertainty drivers and introduce a gradient-based consistency recourse method to determine minimal uncertainty interventions that restore consistency. 
The results show that inconsistency under uncertainty can be effectively learned, analyzed, and repaired through response-surface modeling, providing a scalable foundation for uncertainty-aware consistency management in CPS development.
\end{abstract}

\begin{IEEEkeywords}
Cyber-physical systems, Consistency analysis, Uncertainty quantification, Surrogate model, Algorithmic recourse.
\end{IEEEkeywords}

\section{Introduction}
\label{sec:intro}

Modern Cyber-Physical Systems (CPS) are designed through interacting models describing structural, behavioral, and physical properties of the system~\cite{lee2016introduction, madni2018model}. 
Consistency requires that shared model elements remain compatible across interacting models, and is crucial for reliable system design and operation
~\cite{bhave2011view}. 
At design time, each model is treated as an independent artifact whose admissible state ranges are checked for compatibility with those of interacting models.
However, uncertainty due to parameter variation, sensor noise, and abstraction gaps propagates across interacting models~\cite{troya2021uncertainty, makelburg2026surveying}, potentially inducing inconsistencies at design time that are not observable at nominal operating conditions.

\begin{figure}[t]
    \centering
    \includegraphics[width=0.92\linewidth]{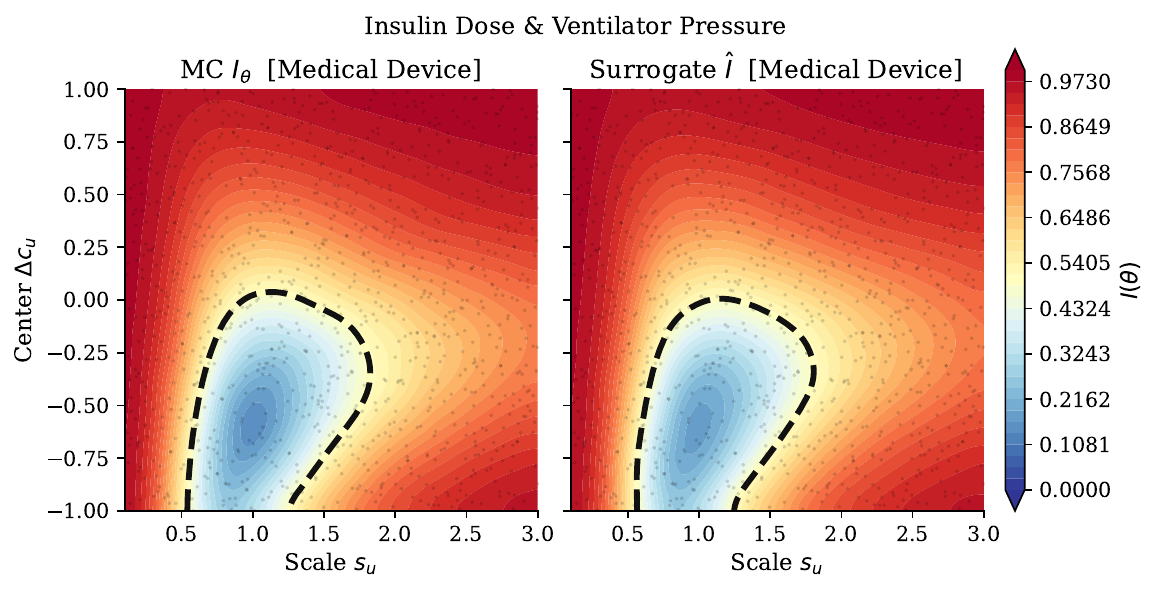}
    \caption{Monte Carlo and surrogate inconsistency response surfaces for the \textit{Medical Device -- Insulin Dose $\&$ Ventilator Pressure} scenario. The surrogate preserves the global response geometry and consistency boundary.}
    \label{fig:landscape_surrogate}
\end{figure}

As an example, consider a development-time medical system for insulin therapy consisting of two interacting (acyclic) models: a metabolic model used to estimate the patients' glucose regulation and a treatment model used to determine appropriate insulin dosages.
Given the current calibrations, both models are consistent as their assumed physiological state ranges overlap.
However, after updating the metabolic model based on new clinical evidence, the feasible state estimates shift systematically, and the feasible state range of the metabolic model shifts and no longer overlaps with the range assumed by the treatment model.
Yet the two models, each individually correct, cannot be jointly satisfied.
Such inconsistencies can propagate across engineering artifacts and can ultimately lead to critical system failures~\cite{mordecai2017minding, nguyen2018cyber}.

When uncertainty is modeled explicitly, consistency becomes a property over sets of admissible model configurations rather than single deterministic values~\cite{scott2016constrained}. 
Reasoning about this property requires propagating uncertainty through interacting models and evaluating large spaces of feasible system states~\cite{althoff2008reachability}.
This results in computationally expensive stochastic response estimation problems similar to those studied in uncertainty quantification, simulation analytics, and surrogate-assisted scientific computing~\cite{sacks1989design, forrester2008engineering, peherstorfer2018survey}. 
Recent complexity results further show that exact containment and overlap reasoning for 
set representations is W[1]-hard with respect to its dimension 
motivating learned approximation and surrogate-based analysis approaches~\cite{froese2026parameterized}.
While existing approaches can determine whether a particular configuration is consistent, they provide limited support for systematically learning and analyzing the global inconsistency behavior induced by uncertainty.

To address these challenges, we reformulate inconsistency analysis under uncertainty as an intervention-response modeling problem, where interventions (like shifting and scaling) parameterize modifications to the underlying uncertainty.
Given an intervention configuration $\theta$, the resulting inconsistency score $I(\theta)$ defines a response surface (representing the probability of inconsistency) over the induced uncertainty space.
This formulation enables systematic generation of intervention-response datasets for scalable response-function estimation, surrogate modeling
, and intervention-based analysis~\cite{peherstorfer2018survey}.
To support scalable exploration of the resulting inconsistency landscapes, we combine multi-fidelity Monte Carlo estimation with a learned surrogate model that directly predicts inconsistency from post-intervention uncertainty geometry.
Figure~\ref{fig:landscape_surrogate} shows the inconsistency landscape for the insulin dose scenario with respect to shift and scaling of the uncertainty region. The learned surrogate models landscape agrees quantitatively and qualitatively with the one estimated via Monte-Carlo sampling.
The surrogate enables efficient approximation of computationally expensive consistency evaluations while preserving geometric dependencies induced by uncertainty propagation across interacting models~\cite{ peherstorfer2016optimal}.
Additionally, it enables large-scale sensitivity analysis and consistency recourse to be tractable.
The learned response surfaces reveal nonlinear interaction effects, dominant uncertainty drivers, and intervention regions that restore cross-model consistency.

Building on this formulation, this paper makes the following contributions:
(i) We formalize inconsistency under uncertainty as an intervention-response function $I(\theta)$ over interacting CPS models.
(ii) We evaluate structured surrogate architectures for learning the inconsistency response surface $I(\theta)$ directly from post-intervention set geometry. 
(iii) We analyze the structural drivers of inconsistency through sensitivity analysis and intervention-response landscapes.
(iv) We address consistency recourse through  population- and gradient-based minimal uncertainty interventions. 

\section{Preliminaries}
\label{sec:preliminaries}
In CPS development, we represent uncertain states as sets of admissible values and analyze how uncertainty propagates across interacting components.

\paragraph{Set-Based Uncertainty Representation}
\label{sec:uncertainty-representation} 
We represent an uncertain model using zonotopes, which compactly encode sets of admissible states via a center point and a set of generator vectors spanning the uncertainty region.
In particular, we use constrained zonotopes~\cite{scott2016constrained}, which allow representing bounded uncertainties and linear dependencies, while supporting efficient algebraic operations such as affine transformations 
~\cite{schafer2023scalable, althoff2007reachability} 
\begin{definition}[Constrained Zonotope \cite{scott2016constrained}] 
\label{def:constrained_zonotope} 
Given a center $c\in \mathbb{R}^{n}$, a generator matrix $G = [g^{(1)}, \cdots, g^{(\gamma)}] \in \mathbb{R}^{n \times \gamma}$ with $\gamma \in \mathbb{N}$ generator vectors, and constraints given by $A \in \mathbb{R}^{n \times \gamma}$, $b \in \mathbb{R}^{n}$, a constrained zonotope $\mathcal{Z}$ is defined as
\begin{align*}
\mathcal{Z} := \Big\{ x \in \mathbb{R}^{n} \, \Big| \, x &= c + \sum_{i = 1}^{\gamma} \xi^{(i)} g^{(i)}, A \xi = b, \, \xi \in [-1,1]^{\gamma} \Big\}
\end{align*}
\end{definition}
Diagonal generators encode interval uncertainty~\cite{girard2005reachability,scott2016constrained}, while covariance-derived generators approximate probabilistic confidence regions~\cite{hardle2007applied}.

\paragraph{Consistency under Uncertainty}
Given a collection of uncertain models, each represented by a constrained zonotope, consistency requires that their uncertainty sets admit at least one jointly feasible realization.
\begin{definition}
Let $\mathcal Z = \{\mathcal Z_1,\dots,\mathcal Z_n\}$ denote a collection of interacting uncertainty sets.
The uncertainty sets are jointly consistent iff there exists at least one jointly feasible realization:
\label{eq:consistency-under-uncertainty}
$\bigcap_{i=1}^{n} \mathcal Z_i \neq \emptyset.$    
\end{definition}

Thus, consistency under uncertainty holds when there exists at least one realization that is admissible across all uncertainty sets simultaneously.

\paragraph{Uncertainty Mappings}
\label{sec:uncertainty_propagation}

Dependencies between uncertainty sets are captured through uncertainty mappings (UMs), which define directed relations between source and target models. 
We assume that these mappings induce a directed acyclic dependency graph. 
As in the insulin therapy example, the physiological state ranges assumed by the metabolic and treatment models must overlap.
\begin{definition}
\label{def:uncertainty-mapping}
A UM is an affine function $\varphi : X_i \rightarrow X_j$, typically of affine form $\varphi(x)=Fx+f$, that relates the admissible states of a source model to those of a target model. 
The mapped uncertainty set is defined as
$\Phi (\mathcal{Z}_j)
=
\varphi(\mathcal{Z}_i)\cap\mathcal{Z}_j. $   
\end{definition}
The intersection checks whether the mapped source states are compatible with the locally admissible uncertainty regions. 
\begin{definition}
\label{def:system-consistency}
The global feasible region is defined as the intersection of all mapped sets:
$\mathcal{Z}_{\mathrm{sys}}
=
\bigcap_i \Phi(\mathcal{Z}_i).$    
\end{definition}

\section{Methodology}
\label{sec:approach}
This section reformulates uncertainty-aware consistency analysis as a response-surface modeling problem over intervention-induced uncertainty configurations.
Intervention configurations $\theta$ define the intervention space, induced inconsistency $I(\theta)$ defines the response surface. 

\subsection{CPS Representation and Inconsistency Measure}
\label{sec:param-CPS}
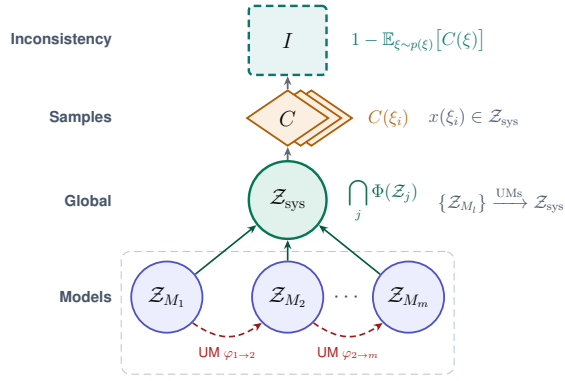
\begin{figure}[t]
\centering
\resizebox{0.85\columnwidth}{!}{%
    \begin{tikzpicture}[
    font=\sffamily,
    >={Stealth[length=6pt,width=5pt]},
    model node/.style={
        circle,
        draw=modelblue!80!black,
        line width=1.2pt,
        fill=modelblue!12,
        minimum size=1.65cm,
        inner sep=3pt,
        font=\Large,
    },
    global node/.style={
        circle,
        draw=globalgreen!80!black,
        line width=1.6pt,
        fill=globalgreen!12,
        minimum size=1.8cm,
        inner sep=3pt,
        font=\Large,
    },
    score node/.style={
        diamond,
        draw=scoregold!80!black,
        line width=1.2pt,
        fill=scoregold!12,
        minimum width=1.9cm,
        minimum height=1.3cm,
        inner sep=3pt,
        font=\Large,
        aspect=1.5,
    },
    consistency node/.style={
        rectangle,
        draw=consteal!80!black,
        line width=1.6pt,
        fill=consteal!10,
        minimum width=1.8cm,
        minimum height=1.6cm,
        inner sep=4pt,
        font=\Large,
        rounded corners=3pt,
        dashed,
        dash pattern=on 5pt off 3pt,
    },
    causal arrow/.style={
        draw=titledark!70,
        line width=1.2pt,
        ->,
    },
    um arrow/.style={
        draw=umarrow!70!black,
        line width=1.1pt,
        ->,
        dashed,
        dash pattern=on 4pt off 2.5pt,
    },
    converge arrow/.style={
        draw=globalgreen!60!black,
        line width=1.1pt,
        ->,
    },
    level box/.style={
        draw=levelbox!60,
        line width=0.7pt,
        dashed,
        dash pattern=on 5pt off 3pt,
        rounded corners=6pt,
        inner sep=6pt,
        fill=none,
    },
    level label/.style={
        font=\normalsize\sffamily\bfseries,
        text=arrowlabel,
        anchor=east,
    },
]

\node[model node] (ZM1) at (-2.8,0)
    {$\mathcal{Z}_{M_1}$};

\node[model node] (ZM2) at (0,0)
    {$\mathcal{Z}_{M_2}$};

\node[model node] (ZM3) at (2.8,0)
    {$\mathcal{Z}_{M_m}$};

\node[font=\Large,text=arrowlabel] at (1.4,0)
    {$\cdots$};

\draw[um arrow]
    (ZM1.south east)
    to[out=-35,in=-145,looseness=1.15]
    node[
        midway,
        below=3pt,
        font=\small\sffamily,
        text=umarrow!80!black
    ] (UM12) {UM $\varphi_{1\to 2}$}
    (ZM2.south west);

\draw[um arrow]
    (ZM2.south east)
    to[out=-35,in=-145,looseness=1.15]
    node[
        midway,
        below=3pt,
        font=\small\sffamily,
        text=umarrow!80!black
    ] (UM2m) {UM $\varphi_{2\to m}$}
    (ZM3.south west);

\begin{scope}[on background layer]
    \node[
        level box,
        fit=(ZM1)(ZM2)(ZM3)(UM12)(UM2m)
    ] (modelbox) {};
\end{scope}

\node[global node] (Zsys) at (0,2.25)
    {$\mathcal{Z}_{\mathrm{sys}}$};

\draw[converge arrow] (ZM1) -- (Zsys);
\draw[converge arrow] (ZM2) -- (Zsys);
\draw[converge arrow] (ZM3) -- (Zsys);

\node[
    font=\large,
    text=globalgreen!70!black,
    anchor=west
] (Zsyslabel) at ([xshift=10pt]Zsys.east)
    {$\displaystyle\bigcap_j\Phi(\mathcal{Z}_j)$};

\node[
    font=\large,
    text=arrowlabel,
    anchor=west
] at ([xshift=6pt]Zsyslabel.east)
    {$\{\mathcal{Z}_{M_l}\}
      \xrightarrow{\mathrm{UMs}}
      \mathcal{Z}_{\mathrm{sys}}$};

\node[score node] at (0.55,4.15) {};
\node[score node] at (0.30,4.15) {};
\node[score node] (C) at (0,4.15)
    {$C$};

\draw[causal arrow] (Zsys) -- (C);

\node[
    font=\large,
    text=scoregold!80!black,
    anchor=west
] (Clabel) at ([xshift=22pt]C.east)
    {$C(\xi_i)$};

\node[
    font=\large,
    text=arrowlabel,
    anchor=west
] at ([xshift=6pt]Clabel.east)
    {$x(\xi_i)\in\mathcal{Z}_{\mathrm{sys}}$};

\node[consistency node] (Itheta) at (0,5.95)
    {$I$};

\draw[causal arrow] (C) -- (Itheta);

\node[
    font=\large,
    text=consteal!80!black,
    anchor=west
] at ([xshift=12pt]Itheta.east)
    {$1-\mathbb{E}_{\xi\sim p(\xi)}
      \bigl[C(\xi)\bigr]$};

\node[level label] at (-3.95,0)
    {Models};

\node[level label] at (-3.95,2.25)
    {Global};

\node[level label] at (-3.95,4.15)
    {Samples};

\node[level label] at (-3.95,5.95)
    {Inconsistency};

\end{tikzpicture}
}
\caption{Consistency reasoning pipeline: uncertainty sets are mapped and intersected into a global feasibility region, sampled to produce binary consistency scores, and aggregated into the inconsistency response $I(\theta)$.}
\label{fig:problem_overview}
\end{figure}

We consider a CPS as a collection of interacting models $\{M_1, \ldots, M_m\}$, each representing a distinct aspect of the system, such as physical behavior, control logic, or sensor characteristics.
Following the set-based uncertainty representation introduced in Definition~\ref{def:constrained_zonotope},
each model $M_l$ is represented by a constrained zonotope $\mathcal{Z}_{M_l} \subseteq \mathbb{R}^{n}$ representing the set of admissible states of the model under uncertainty.
The dimension $n$ represents the number of independent uncertainties affecting $M_l$. 
Dependencies between models are captured through UMs (Definition~\ref{def:uncertainty-mapping}), and their intersection determines the global consistency region $\mathcal{Z}_\mathrm{sys}$ (Definition~\ref{def:system-consistency}). 
Figure~\ref{fig:problem_overview} provides an overview of the consistency evaluation pipeline.

The consistency of a CPS is evaluated in two steps:
First, model-level uncertainty sets are propagated through uncertainty mappings to determine the global feasible region $\mathcal{Z}_{\mathrm{sys}}$.
This region contains all states that remain jointly admissible across the interacting model after accounting for all dependencies induced by the uncertainty mappings.

Given a sampled realization $\xi$ 
the resulting state $x(\xi)$ is evaluated against the global feasible region $\mathcal{Z}_{\mathrm{sys}}$ to determine whether the system is consistent. 
The binary consistency score is defined as:
\begin{equation}
C(\xi) = \begin{cases}
1, & x(\xi) \in \mathcal{Z}_\mathrm{sys}\\
0, & \text{otherwise}
\end{cases}
\end{equation}
Because $C(\xi)$ depends on stochastic realizations $\xi$, we aggregate over $\xi$ and directly define the inconsistency measure function
\begin{equation}
\label{eq:inconsistency-function}
I = 1 - \mathbb{E}_{\xi \sim p(\xi)}\bigl[C(\xi)\bigr].
\end{equation}

The function $I$ represents the probability that the system becomes inconsistent. 
Evaluating $I$ exactly requires reasoning over the full feasibility region $\mathcal{Z}_\mathrm{sys}$, which is intractable for general CPS configurations due to the computational hardness of exact zonotope  containment~\cite{froese2026parameterized}.
We therefore estimate it numerically.

\subsection{Estimating the Inconsistency Response}
\label{sec:inconsistency-estimation-methods}
We estimate $I$ using three estimators with different accuracy-runtime tradeoffs: 
Monte Carlo sampling (MC), 
a geometric proxy based on axis-aligned bounding boxes (AABB), 
and a multi-fidelity control-variate estimator (MFMC).

\subsubsection{Monte Carlo (MC)}
MC estimates $I$ by averaging binary inconsistency outcomes over $N_{MC}$ sampled realizations $\xi^{(k)} \sim p(\xi)$:
\begin{equation}
\widehat{I}_{\text{MC}} = 1 - \frac{1}{N_{MC}} \sum_{k=1}^{N_{MC}} C(\xi^{(k)}).
\end{equation}
By the law of large numbers, $\widehat{I}_{\text{MC}}$ converges to $I$ as defined in Equation~\ref{eq:inconsistency-function}, making it the most principled estimator.
It is unbiased but computationally expensive~\cite{rubinstein2016simulation} as accurate estimates require a large $N$. \added[id=TS]{ The standard error is $\sqrt{\widehat{I}(1-\widehat{I})/N_{MC}}$, which upper-bounds at $\approx 0.016$ (at I=0.5) for $N_{MC} = 1000$. This sets a label-noise floor on any surrogate fit to MC labels. Note that each $C(\xi)$ is evaluated using an exact linear program membership test.}
\subsubsection{Geometric Proxy AABB} 
A computationally cheaper approximation replaces each zonotope $\mathcal{
Z}_{M_l}$ by its axis-aligned bounding boxes (AABB) $B_l$.
These AABBs are strict over-approximations that enclose the respective zonotopes but ignore their orientations and generator structures.
This converts the exact set intersection into a box overlap problem, which can be evaluated efficiently
using a geometric Jaccard overlap:
\begin{equation}
J_{\text{AABB}} =
\frac{\mathrm{vol}(B_\cap)}
{\mathrm{vol}(B_\cup)}, \quad
B_\cap = \bigcap_i B_i, \quad
B_\cup = \bigcup_i B_i.
\end{equation}
The resulting inconsistency approximation is then $\widehat{I}_{\text{AABB}} = 1 - J_{\text{AABB}}$.
Because bounding boxes extend in all axis directions regardless of zonotope orientation, the union $B_\cup$ grows faster than the intersection $B_\cap$, reducing the Jaccard ratio and causing $I_{\text{AABB}}$ to systematically overestimate inconsistency~\cite{ericson2004real}.

\subsubsection{Multi-Fidelity Estimation (MFMC)}
MFMC reduces the cost of MC estimation by pairing a few-sample MC estimate with AABB as a low-fidelity control variate~\cite{peherstorfer2016optimal}. 
Since both $I_{\text{MC}}$ and $I_{\text{AABB}}$ quantify inconsistency for the same random samples, they are correlated. 
This correlation can be exploited to reduce estimator variance through the correction. 
The resulting estimator is defined as
\begin{equation}
\widehat{I}_{\text{MF}} = \widehat{I}_{\text{MC}} 
    + \alpha \cdot \bigl( \mu_{\text{AABB}} - \widehat{I}_{\text{AABB}}\bigr),
\end{equation}

where $\widehat{I}_{\text{AABB}}$ denotes the AABB inconsistency estimate on the same sample as $\widehat{I}_{\text{MC}}$, $\mu_{\text{AABB}}$ 
is a large-sample AABB estimate (which is cheap to obtain), and $\alpha
= \frac{\mathrm{Cov}(I_{\text{MC}},I_{\text{AABB}})}
{\mathrm{Var}(I_{\text{AABB}})}$ is the optimal control-variate coefficient estimated from the same set~\cite{peherstorfer2016optimal}. \added[id=TS]{Because of that, MFMC has a $\mathcal{O}(1/N)$ finite-sample bias. In our experiments, $r(\widehat{I}_{MC},\widehat{I}_{AABB})\approx 0.94$, so the achieved variance-reduction $1- r^2 \approx 0.09-0.13$, leading to an 8-11x variance reduction.}

\subsection{Dataset Construction}
\label{sec:dataset-construction}
To systematically analyze how uncertainty shapes inconsistency, we introduce an intervention configuration $\theta \in \Theta \subseteq \mathbb{R}^3$ that parameterizes the uncertainty geometry of each model $M_l$.
This induces parameterized sets $\mathcal{Z}_{M_l}(\theta)$, a global feasibility region $\mathcal{Z}_\mathrm{sys}(\theta)$,
and transforms the inconsistency response into a function 
\begin{equation}
\label{eq:inconsistency-response-function}
I(\theta) = 1 - \mathbb{E}_{\xi \sim p(\xi)}[C(\xi, \theta)]
\end{equation}
over the intervention space $\Theta$.

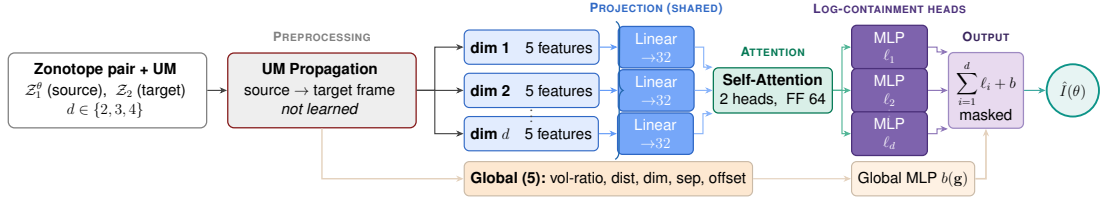
\begin{figure*}[t]
\centering
\resizebox{0.8\linewidth}{!}{%
    \begin{tikzpicture}[
  font=\sffamily\large,
  >={Stealth[length=6pt,width=5pt]},
  inputbox/.style={
    draw=black!50, fill=white, rounded corners=5pt,
    line width=1.0pt, inner sep=7pt, align=center
  },
  umbox/.style={
    draw=umarrow!60!black, fill=uprGray, rounded corners=5pt,
    line width=1.0pt, inner sep=7pt, align=center
  },
  dimbox/.style={
    draw=dimBlue!80!black, fill=dimBlue!15, rounded corners=4pt,
    line width=0.9pt, inner sep=5pt, align=center,
    minimum width=3.0cm, minimum height=0.95cm
  },
  globbox/.style={
    draw=globOrg!80!black, fill=globOrg, rounded corners=4pt,
    line width=0.9pt, inner sep=5pt, align=center,
    minimum height=0.95cm
  },
  attnbox/.style={
    draw=attnGreen!80!black, fill=attnGreen!15, rounded corners=4pt,
    line width=1.0pt, inner sep=5pt, align=center,
    minimum width=2.5cm, minimum height=1.0cm
  },
  logbox/.style={
    draw=logPlum!80!black, fill=logPlum!80, rounded corners=4pt,
    line width=1.0pt, inner sep=5pt, align=center,
    minimum width=2.1cm, minimum height=0.95cm,
    text=white
  },
  globmlpbox/.style={
    draw=globOrg!80!black, fill=globOrg!60, rounded corners=4pt,
    line width=0.9pt, inner sep=5pt, align=center,
    minimum width=2.1cm, minimum height=0.95cm
  },
  sumbox/.style={
    draw=logPlum!60, fill=concatPurp, rounded corners=4pt,
    line width=1.0pt, inner sep=6pt, align=center,
    minimum width=1.6cm
  },
  outnode/.style={
    circle, draw=consteal, fill=consteal!10, line width=1.5pt,
    inner sep=4pt, align=center, minimum size=1.45cm
  },
  arr/.style={draw=arrowGray, line width=0.9pt, ->},
  arrbold/.style={draw=arrowGray!80, line width=1.3pt, ->},
  seclab/.style={font=\sffamily\normalsize\bfseries, text=black!50},
]

\node[inputbox] (inp) at (0, 0) {%
  \begin{tabular}{c}
    \textbf{Zonotope pair + UM}\\[2pt]
    $\mathcal{Z}_1^\theta$ (source),\;
    $\mathcal{Z}_2$ (target)\\[1pt]
    $d \in \{2,3,4\}$
  \end{tabular}
};

\node[umbox, right=0.55cm of inp] (um) {%
  \begin{tabular}{c}
    \textbf{UM Propagation}\\[2pt]
    source $\to$ target frame\\[1pt]
    \textit{not learned}
  \end{tabular}
};
\draw[arr] (inp) -- (um);

\node[dimbox, right=1.3cm of um, yshift=1.2cm]  (dim1) {%
  \textbf{dim 1}\quad 5 features};
\node[dimbox, right=1.3cm of um]                (dim2) {%
  \textbf{dim 2}\quad 5 features};
\node[dimbox, right=1.3cm of um, yshift=-1.2cm] (dimd) {%
  \textbf{dim $d$}\quad 5 features};

\node[font=\large, text=arrowGray]
  at ($(dim2.south)!0.5!(dimd.north)$) {$\vdots$};

\draw[arr] (um.east) -- ++(0.5,0) |- (dim1.west);
\draw[arr] (um.east) -- ++(0.5,0) |- (dim2.west);
\draw[arr] (um.east) -- ++(0.5,0) |- (dimd.west);

\node[globbox, right=1.3cm of um, yshift=-2.45cm] (glob) {%
  \textbf{Global (5):} vol-ratio, dist, dim, sep, offset};
\draw[arr, draw=globOrg!90!black, line width=1.1pt]
  (um.south) |- (glob.west);

\node[logbox, right=0.5cm of dim1, fill=dimBlue!70, draw=dimBlue!80!black]
  (proj1) {Linear\\${\to}32$};
\node[logbox, right=0.5cm of dim2, fill=dimBlue!70, draw=dimBlue!80!black]
  (proj2) {Linear\\${\to}32$};
\node[logbox, right=0.5cm of dimd, fill=dimBlue!70, draw=dimBlue!80!black]
  (projd) {Linear\\${\to}32$};

\foreach \d/\p in {dim1/proj1, dim2/proj2, dimd/projd}{
  \draw[arr, draw=dimBlue!80] (\d.east) -- (\p.west);
}

\draw[decorate, decoration={brace, amplitude=6pt, mirror},
      draw=bracketCol, line width=1.0pt]
  ($(projd.south west)+(-0.05,-0.12)$) --
  ($(proj1.north west)+(-0.05, 0.12)$);

\node[attnbox, right=0.5cm of proj2] (attn) {%
  \textbf{Self-Attention}\\[1pt]
  2 heads,\; FF 64
};

\draw[arr, draw=dimBlue!60] (proj1.east) -- ++(0.25,0) |- (attn.north west);
\draw[arr, draw=dimBlue!60] (proj2.east) -- (attn.west);
\draw[arr, draw=dimBlue!60] (projd.east) -- ++(0.25,0) |- (attn.south west);

\node[logbox, right=0.5cm of attn, yshift=1.2cm]  (lc1) {MLP\\$\ell_1$};
\node[logbox, right=0.5cm of attn]                 (lc2) {MLP\\$\ell_2$};
\node[logbox, right=0.5cm of attn, yshift=-1.2cm]  (lcd) {MLP\\$\ell_d$};

\node[font=\large, text=logPlum!60]
  at ($(lc2.south)!0.5!(lcd.north)$) {$\vdots$};

\draw[arr, draw=attnGreen!70] (attn.east) -- ++(0.25,0) |- (lc1.west);
\draw[arr, draw=attnGreen!70] (attn.east) -- (lc2.west);
\draw[arr, draw=attnGreen!70] (attn.east) -- ++(0.25,0) |- (lcd.west);

\node[globmlpbox, right=0.5cm of attn, yshift=-2.45cm] (gmlp)
  {Global MLP $b(\mathbf{g})$};
\draw[arr, draw=globOrg!90!black] (glob.east) -- ++(0.25,0) |- (gmlp.west);

\node[sumbox, right=0.55cm of lc2] (sumnode) {%
  $\displaystyle\sum_{i=1}^{d}\ell_i + b$\\[2pt]
  masked
};

\draw[arr, draw=logPlum!70] (lc1.east) -- ++(0.25,0) |- (sumnode.north west);
\draw[arr, draw=logPlum!70] (lc2.east) -- (sumnode.west);
\draw[arr, draw=logPlum!70] (lcd.east) -- ++(0.25,0) |- (sumnode.south west);
\draw[arrbold, draw=globOrg!90!black]
  (gmlp.east) -| (sumnode.south);

\node[outnode, right=0.6cm of sumnode] (out) {$\hat{I}(\theta)$};
\draw[arr, draw=consteal!80] (sumnode.east) -- (out.west);

\node[seclab, above=0.18cm of um] {\textsc{Preprocessing}};
\node[seclab, text=dimBlue!80!black, above=0.18cm of proj1]
  {\textsc{Projection (shared)}};
\node[seclab, text=attnGreen!80!black, above=0.18cm of attn]
  {\textsc{Attention}};
\node[seclab, text=logPlum!80!black, above=0.18cm of lc1]
  {\textsc{Log-containment heads}};
\node[seclab, text=logPlum!60!black, above=0.18cm of sumnode]
  {\textsc{Output}};

\end{tikzpicture}
}
\caption{Product-Set Transformer (PST) surrogate. Per-dimension features are projected, processed by a  self-attention block, and decoded into per-dimension log-containment estimates whose sum encodes the conjunction  structure of zonotope overlap.}
\label{fig:surrogate_model}
\end{figure*}

The estimators above can evaluate $I(\theta)$ for a fixed $\theta$, but large-scale analysis requires a dense coverage of $\Theta$.
We therefore construct an intervention-response dataset that serves two purposes:
enabling the training of a surrogate model for efficient evaluation, and supporting downstream sensitivity analysis and recourse search.

$N$ configurations $\theta_i \in \Theta$ are samples using Saltelli sampling~\cite{saltelli2002making} to enable variance-based sensitivity analysis.
Each configuration is labeled using MC, resulting in the dataset:
\begin{equation}
\label{eq:dataset}
    D = \{ (\theta_i, \hat{I}(\theta_i)) \}_{i=1}^{N},
\end{equation}

The resulting dataset $D$ provides a sample approximation of the inconsistency response landscape over $\Theta$.

\subsection{Learning Inconsistency Response Surfaces}

The problem with the previously mentioned approaches, especially MC, is that the estimation of $I$ is expensive, and can quickly become intractable for problems requiring many evaluations of $I$. In fact, Froese et al.~\cite{froese2026parameterized} show that containment of a zonotope within another (which is essential to the calculation of $I$) is coNP-complete and W$[1]$-hard in the dimension $d$. They further show a duality between ReLU networks and zonotopes, which motivates our surrogate model with ReLU nonlinearities.

Our proposed surrogate model takes as input a pair of zonotopes, $(\mathcal{Z}_{1},\mathcal{Z}_{2})$, representing the source and target zonotopes. It then approximates the probability of inconsistency $\Hat{I}$, i.e.
\begin{equation}
\hat{I} \approx 1 - \Pr\!\left(c_1 + G_1\xi \in \mathcal{Z}_2\right),
\qquad \xi \sim \mathcal{U}([-1,1]^\gamma),\end{equation}
The features to represent the zonotopes are computed in the propagated source frame (i.e., after applying the UMs), and are relative. This makes the representation invariant to a common translation, rotation, or scaling of the two zonotopes, which is the desired inductive bias for the above probability and enables the model to generalize to zonotopes of different scales.

\paragraph{Per-Dimension Features} We represent relative differences between the zonotopes per dimension using five different features. Let $\sigma_i=r_{1,i}+r_{2,i}$ (where $r_{k,i}=||G_{k,i:}||_2$), then we define the following features: the normalized center offset $(c_1-c_2)_i/\sigma_i$, the source and target per-axis magnitudes $r_{1,i}/\sigma_i,r_{2,i}/\sigma_i$, the generator-row alignment $\cos \angle (G_{1,i},G_{2,i})$, and finally the width-ratio $r_{1,i}/r_{2,i}$.

\paragraph{Global Features} The per-dimension features cannot capture all aspects of alignment between the zonotopes. Hence, we add additional global features: the log-volume ratio ($\sum_{i=1}^d\log r_{1,i}/r_{2,i}$), normalized center distance ($||c_1-c_2||/(\text{avg}_i(\sigma_i)+\epsilon)$), dimension $d \in \{2,3,4\}$, and two support function features: $||c_1-c_2||/(h_1+h_2), ||c_1-c_2||/h_2$. $h_k=\sum_j|G_{k,j}^\top v|$ is the support along the offset direction $v=(c_1-c_2)/||c_1-c_2||$. These two features evaluate the agreement of the zonotopes along their offset vector instead of the coordinate axes. The rationale is that two zonotopes can be inconsistent even for a small offset if they are both narrow along $v$.
\arxivonly{The surrogate uses the explicit geometry of both models, supporting analysis of known configurations rather than prediction for unspecified designs.}

\paragraph{Product-Set Transformer}
To transform the features into an inconsistency estimate, we use a \emph{product-set transformer}. An overview of the architecture is shown in Figure~\ref{fig:surrogate_model}. We first project all per-dimension feature vectors to a 32-dim embedding. Then, we apply self-attention (2 heads, 64-dim feed-forward) to allow explicit interaction between the dimensions. A final per-dimension MLP outputs a logit $l_i$. The final output is then the joint non-containment probability:
\begin{equation}
    \Hat{I} = 1 - \prod_i p_i \cdot p_g = 1 - \exp(\sum_i \log p_i + \log p_g).
\end{equation}
Here, the per-dimension and global logits are transformed to probabilities using the sigmoid function $s$: $p_i=s(l_i), p_g=s(MLP(g))$, where $g$ is the global feature vector. This architecture models a noisy AND: for containment, a point must lie in both zonotopes in all dimensions. The intuition of the architecture is as follows.
For diagonal generators, the joint probability factorizes into a product of per-dimension containment probabilities. But for non-diagonal generators, the probabilities of containment are statistically dependent between dimensions. Because self-attention is applied before the per-dimension factors, each factor $p_i$ is \replaced[id=JM]{conditioned on all other dimensions via attention, so their product implements an attention-parameterized factorization of the joint non-containment probability.}{conditioned on all other dimensions, so their product represents a chain-rule-style factorization of joint containment.}
\paragraph{Training}
We train two instances of the surrogate model. One model (PST-2D3D) for two- and three-dimensional scenarios and one model (PST-4D) for four-dimensional scenarios. We train the models for 150 epochs on a mixture of real and synthetic (for 3D) zonotope scenarios using the Huber Loss and the AdamW optimizer with a learning rate of $10^{-3}$. For the ground-truth labels, we use the inconsistency calculated using the full MC scheme.

\subsection{Analyzing Inconsistency Response Surfaces}
To systematically analyze the response surface $\Theta$, we parametrize the modification of the uncertainty representation through an intervention vector.  

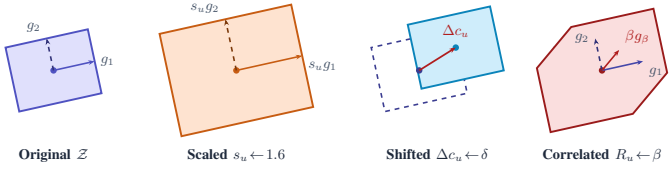
\begin{figure}[t]
\centering
\resizebox{\columnwidth}{!}{%
    \begin{tikzpicture}[font=\sffamily\small, >={Stealth[length=4pt, width=3pt]}]

\def\sep{4.0}
\def\lbl{\small}

\begin{scope}[xshift=0cm]
  \coordinate (C)  at (0,0);
  \coordinate (G1) at (0.9,0.2);
  \coordinate (G2) at (-0.15,0.7);
  \coordinate (P1) at ($(C)+(G1)+(G2)$);
  \coordinate (P2) at ($(C)+(G1)-(G2)$);
  \coordinate (P3) at ($(C)-(G1)-(G2)$);
  \coordinate (P4) at ($(C)-(G1)+(G2)$);
  \fill[modelblue!20, draw=modelblue!80!black, line width=1.2pt]
      (P1)--(P2)--(P3)--(P4)--cycle;
  \fill[modelblue!80!black] (C) circle (2pt);
  \draw[->, modelblue!80!black, line width=1pt] (C) -- ($(C)+(G1)$)
      node[right, font=\lbl, text=arrowlabel] {$g_1$};
  \draw[->, modelblue!50!black, line width=1pt, dashed] (C) -- ($(C)+(G2)$)
      node[above left, font=\lbl, text=arrowlabel] {$g_2$};
  \node[font=\small\bfseries, text=titledark, anchor=north] at (0,-1.6)
      {Original $\mathcal{Z}$};
\end{scope}

\begin{scope}[xshift=\sep cm]
  \coordinate (C)   at (0,0);
  \coordinate (G1)  at (0.9,0.2);
  \coordinate (G2)  at (-0.15,0.7);
  \coordinate (sG1) at (1.44,0.32);
  \coordinate (sG2) at (-0.24,1.12);

  \coordinate (oP1) at ($(C)+(G1)+(G2)$);
  \coordinate (oP2) at ($(C)+(G1)-(G2)$);
  \coordinate (oP3) at ($(C)-(G1)-(G2)$);
  \coordinate (oP4) at ($(C)-(G1)+(G2)$);
  \draw[modelblue!60!black, line width=1pt, dashed]
      (oP1)--(oP2)--(oP3)--(oP4)--cycle;

  \coordinate (P1) at ($(C)+(sG1)+(sG2)$);
  \coordinate (P2) at ($(C)+(sG1)-(sG2)$);
  \coordinate (P3) at ($(C)-(sG1)-(sG2)$);
  \coordinate (P4) at ($(C)-(sG1)+(sG2)$);
  \fill[outputcol!20, draw=outputcol!80!black, line width=1.2pt]
      (P1)--(P2)--(P3)--(P4)--cycle;
  \fill[outputcol!80!black] (C) circle (2pt);
  \draw[->, outputcol!80!black, line width=1pt] (C) -- ($(C)+(sG1)$)
      node[below right, font=\lbl, text=arrowlabel] {$s_u g_1$};
  \draw[->, outputcol!50!black, line width=1pt, dashed] (C) -- ($(C)+(sG2)$)
      node[above left, font=\lbl, text=arrowlabel] {$s_u g_2$};
  \node[font=\small\bfseries, text=titledark, anchor=north] at (0,-1.6)
      {Scaled $s_u \!\leftarrow\! 1.6$};
\end{scope}

\begin{scope}[xshift=2*\sep cm]
  \coordinate (C0) at (0,0);
  \coordinate (C1) at (0.8,0.5);
  \coordinate (G1) at (0.9,0.2);
  \coordinate (G2) at (-0.15,0.7);

  \coordinate (oP1) at ($(C0)+(G1)+(G2)$);
  \coordinate (oP2) at ($(C0)+(G1)-(G2)$);
  \coordinate (oP3) at ($(C0)-(G1)-(G2)$);
  \coordinate (oP4) at ($(C0)-(G1)+(G2)$);
  \draw[modelblue!60!black, line width=1pt, dashed]
      (oP1)--(oP2)--(oP3)--(oP4)--cycle;

  \coordinate (P1) at ($(C1)+(G1)+(G2)$);
  \coordinate (P2) at ($(C1)+(G1)-(G2)$);
  \coordinate (P3) at ($(C1)-(G1)-(G2)$);
  \coordinate (P4) at ($(C1)-(G1)+(G2)$);
  \fill[gnnbox!20, draw=gnnbox!80!black, line width=1.2pt]
      (P1)--(P2)--(P3)--(P4)--cycle;
  \fill[modelblue!60!black] (C0) circle (2pt);
  \fill[gnnbox!80!black]    (C1) circle (2pt);
  \draw[->, uprarrow!80!black, line width=1.1pt]
      (C0) -- (C1)
      node[above, font=\lbl, text=uprarrow!80!black, yshift=3pt]
      {$\Delta c_u$};
  \node[font=\small\bfseries, text=titledark, anchor=north] at (0.4,-1.6)
      {Shifted $\Delta c_u \!\leftarrow\! \delta$};
\end{scope}

\begin{scope}[xshift=3*\sep cm]
  \coordinate (C)  at (0,0);
  \coordinate (G1) at (0.9,0.2);
  \coordinate (G2) at (-0.15,0.7);
  \coordinate (Gb) at (0.375,0.45);

  \coordinate (oP1) at ($(C)+(G1)+(G2)$);
  \coordinate (oP2) at ($(C)+(G1)-(G2)$);
  \coordinate (oP3) at ($(C)-(G1)-(G2)$);
  \coordinate (oP4) at ($(C)-(G1)+(G2)$);
  \draw[modelblue!60!black, line width=1pt, dashed]
      (oP1)--(oP2)--(oP3)--(oP4)--cycle;

  \coordinate (Q1) at ($(C)+(G1)+(G2)+(Gb)$);
  \coordinate (Q2) at ($(C)+(G1)-(G2)+(Gb)$);
  \coordinate (Q3) at ($(C)+(G1)-(G2)-(Gb)$);
  \coordinate (Q4) at ($(C)-(G1)-(G2)-(Gb)$);
  \coordinate (Q5) at ($(C)-(G1)+(G2)-(Gb)$);
  \coordinate (Q6) at ($(C)-(G1)+(G2)+(Gb)$);
  \fill[uprarrow!15, draw=uprarrow!70!black, line width=1.2pt]
      (Q1)--(Q2)--(Q3)--(Q4)--(Q5)--(Q6)--cycle;
  \fill[uprarrow!70!black] (C) circle (2pt);
  \draw[->, modelblue!70!black, line width=1pt] (C) -- ($(C)+(G1)$)
      node[below right, font=\lbl, text=arrowlabel] {$g_1$};
  \draw[->, modelblue!50!black, line width=1pt, dashed] (C) -- ($(C)+(G2)$)
      node[left, font=\lbl, text=arrowlabel] {$g_2$};
  \draw[->, uprarrow!80!black, line width=1pt] (C) -- ($(C)+(Gb)$)
      node[above right, font=\lbl, text=uprarrow!80!black] {$\beta g_\beta$};
  \node[font=\small\bfseries, text=titledark, anchor=north] at (0,-1.6)
      {Correlated $R_u \!\leftarrow\! \beta$};
\end{scope}

\end{tikzpicture}
}
\caption{Intervention types on a zonotope (pre-intervention on the left): scaling, center shift, and correlation.}
\label{fig:uncertainty-parameter}
\end{figure}

%
We define an intervention vector as $\theta = [s_u, \Delta c_u, R_u]^T$,
where $s_u$ parameterizes scale, $\Delta c_u$ center-shift, and $R_u$ correlation interventions, respectively.
Figure~\ref{fig:uncertainty-parameter} illustrates the three types of interventions.
For each uncertainty $u$, we define $c_u(\theta) = c_u + \Delta c_u, \quad G_u(\theta) = s_u G_u R_u$.
Generator coefficients $\xi_i$ represent stochastic realizations of the uncertainty sets and are sampled during consistency evaluation.
We apply an intervention as a controlled perturbation of the analysis parameter, written $\theta \leftarrow w$; this denotes a what-if assignment to the uncertainty model.
\paragraph{Variance-Based Structural Sensitivity Analysis}
To quantify the relative importance of uncertainty factors and their interactions, we employ Sobol variance decomposition. 
The first-order Sobol index is defined as 
\begin{equation}
    S_j = 
    \frac{\mathrm{Var}_{\theta_j} \left( \mathbb{E}_{\bm{\theta}_{\sim j}} [I(\bm{\theta}) \mid \theta_j]\right)}
    {\mathrm{Var}(I(\bm{\theta}))}
\end{equation} 
quantifying the contribution of parameter $\theta_j$ alone to the variance of the inconsistency function $I(\bm{\theta})$.

The total-effect index 
\begin{equation}
    S_j^T= 
    1 -
    \frac{
    \mathrm{Var}_{\bm{\theta}_{\sim j}} \left(
\mathbb{E}[I(\bm{\theta}) \mid \bm{\theta}_{\sim j}]
\right)
    }{\mathrm{Var}(I(\bm{\theta}))}
\end{equation}
captures both isolated and interaction-driven variance contributions involving $\theta_j$.
\subsection{Consistency Recourse}
\label{sec:counterfactual}
While intervention effects help to identify which uncertainty factors contribute the most to inconsistency, they do not indicate how consistency can be repaired.
We therefore investigate how inconsistent uncertainty configurations can be repaired, while keeping the repaired parameters close to the original ones.
Given an inconsistent intervention configuration $\theta^\ast$ with $I > l$, where $l$ denotes the operational consistency threshold, the goal is to identify an intervention restoring consistency with a small change in $\theta^\ast$. 

Formally, we search for a recourse intervention
\begin{equation}
\label{eq:optim-goal}
    \theta'
=
\arg\min_{\theta \in \Theta}
\|\theta-\theta^\ast\|_2
\quad
\text{s.t.}
\quad
\hat I(\theta)\leq l,
\end{equation}

where distances are computed in the normalized intervention space.

This amounts to a continuous optimization problem, and we turn the constrained objective into an unconstrained objective with a penalty term:
\[
L(\theta)
=
\|\theta-\theta^\ast\|_2^2
+
\lambda \cdot
\max(0,\hat I(\theta)-l)^2.
\]

We approach this optimization using both the MFMC estimator (within an evolutionary search algorithm (CMA-ES) and using finite-difference approximated gradients) and the surrogate model (which allows for direct gradient-based search since the surrogate model is fully differentiable w.r.t. $\theta$).

%


The repair vector $\delta\theta = \theta' - \theta^\ast$ quantifies which uncertainty parameters require modification to restore consistency. 

\section{Experimental Analysis}
\label{sec:evaluation}

This section analyzes the inconsistency response function $I(\theta)$ across heterogeneous CPS domains using the dataset construction and estimation pipeline introduced in Section~\ref{sec:dataset-construction}. 
We first characterize the resulting inconsistency landscapes across domains before evaluating how reliably $I(\theta)$ can be estimated, how efficiently these landscapes can be explored, and which uncertainty interventions most strongly influence inconsistency behavior.
The analysis addresses the following research questions:

\begin{enumerate}
    \item[RQ1]  How accurately and efficiently can the inconsistency response function $I(\theta)$ be estimated using geometric, multi-fidelity, and learned estimators? (\S\ref{sec:estimator_comparison})
    
    \item[RQ2] Can a learned surrogate model accurately approximate $I(\theta)$ from generated intervention-response datasets while generalizing across unseen CPS scenarios \deleted[id=JM]{and domains}? (\S\ref{sec:surrogate_model})

    \item[RQ3] Which uncertainty parameters and parameter interactions most strongly drive inconsistency behavior across CPS domains? (\S\ref{sec:sensitivity_analyis_results})

    \item[RQ4] Given a configuration that leads to inconsistency, what parameter intervention restores consistency? (\S\ref{sec:counterfactual-eval})
\end{enumerate}

\subsection{Experimental Case Studies}
\label{sec:datasets}
We construct a structured intervention-response dataset spanning 48 scenarios across ten CPS domains. 
The scenarios cover systems engineering inconsistency situations as well as domain-specific CPS applications, including automotive systems, HVAC, industrial robotics, medical devices, railway systems, satellite systems, smart grids, water/chemical processes, and wind turbines. 
The dataset includes interval and probabilistic uncertainty sources derived from publicly available engineering datasets and standards, like 
PLEIAData~\cite{ibarra2023pleiadata}, 
SWaT~\cite{goh2016dataset}, and 
OpenFAST~\cite{jonkman2024openfast}. 

Each scenario defines an intervention space $\theta$ over zonotope-based uncertainty representations and is associated with an inconsistency response $I(\theta)$ as defined in Equation~\ref{eq:inconsistency-response-function}. 
The intervention parameters control the scale, center displacement, and correlation structure of the uncertainty sets, corresponding to their main geometric degrees of freedom. 
The scale factor $s_u \in [0.1, 5.0]$ scales the uncertainty region compared to its nominal size, where $s_u <1$ generates an under approximation, and $s_u > 1 $ an over approximation. 
The normalized center shift $\Delta c_u \in [-1.0, 1.0]$ enables comparable interventions across scenarios with different uncertainty magnitudes. 
The correlation parameter $\rho_u \in [0.0, 0.95]$ ranges from independent to strongly coupled uncertainty dimensions while avoiding degenerate configurations. 
Intervention configurations are sampled using the Saltelli scheme~\cite{saltelli2010variance}, which systematically explores the intervention space $\theta$ by varying the intervention parameters over their respective ranges.
Using $N=2048$ base samples, leads to $16{,}384$ evaluations across the three intervention parameters.
\added[id=JM]{For all MC-based estimates, we use $N_{MC} = 1000$ samples per evaluation and set the consistency threshold $l = 0.5$}
, i.e. inconsistent when most sampled realizations are not jointly feasible.
\added[id=TS]{All estimators are evaluated with an inductive, scenario-level hold-out, i.e. the surrogate (and the other estimators, although they are not trained) is evaluated only on scenarios it has not seen during training in all experiments below.}
\added[id=JM]{Of the 48 scenarios, 35 are used for surrogate training, and 13 are held out for evaluation at the scenario level; held-out scenarios share domains with training scenarios but use distinct zonotope parameterizations and intervention configurations, evaluating within-domain generalization to unseen system configurations.}

Additional implementation details, scenario specifications, dataset resources, and source code are available in the GitHub repository\footnote{\label{fn:repo}\faGithub\url{https://github.com/DE-TUM/IRIS-CPS}}.
\newcounter{fnrepo}
\setcounter{fnrepo}{\value{footnote}}
All experiments were run on an Ubuntu server (AMD EPYC 9224, 24c/48t, 7 TiB SATA SSD), run in a Docker container limited to 16 vCPUs and 250 GiB RAM.

\subsection{RQ1: Estimating the Inconsistency Response Function}
\label{sec:estimator_comparison}

We evaluate how accurately and efficiently the inconsistency response function $I(\theta)$ can be estimated using geometric (AABB), multi-fidelity (MFMC), and learned estimators.

\begin{figure}[t]
    \centering
    \includegraphics[width=0.99\linewidth]{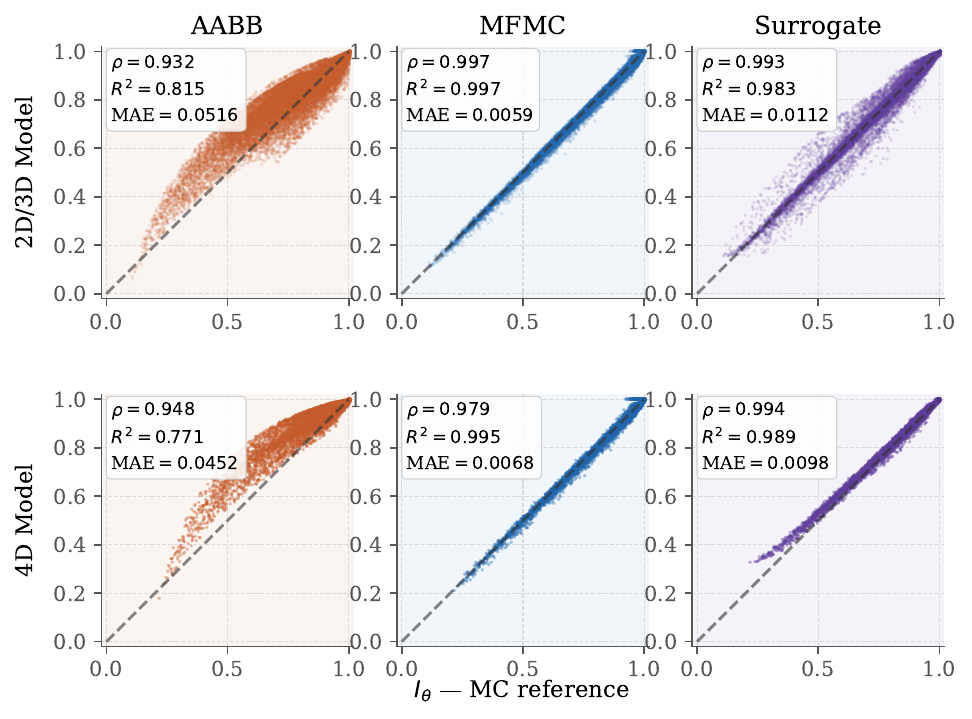}
    \caption{Agreement of AABB, MFMC, and learned surrogate inconsistency estimates with the Monte Carlo reference. MFMC and the learned surrogate achieve substantially lower error and bias than AABB.}
    \label{fig:accuracy_scatter}
\end{figure} 

Figure~\ref{fig:accuracy_scatter} compares AABB, MFMC, and the learned surrogate against the MC reference. 
AABB shows substantially larger deviations and systematic overestimations, particularly in low-inconsistency regions ($\rho=0.932$, $R^2=0.815$).
MFMC closely matches the MC reference ($\rho=0.997$, $R^2=0.997$).
The learned surrogate model also preserves the overall inconsistency structure well ($\rho=0.993$, $R^2=0.983$), and is on par with MFMC in terms of rank correlation for 4D scenarios ($+0.015$).
\added[id = JM]{At $N_{MC}$ = 1000, the MC noise floor is $0.016$, with an MAE of $0.011$ the surrogate falls below it.}

\begin{figure}[t]
    \centering
    \includegraphics[width=0.99\linewidth]{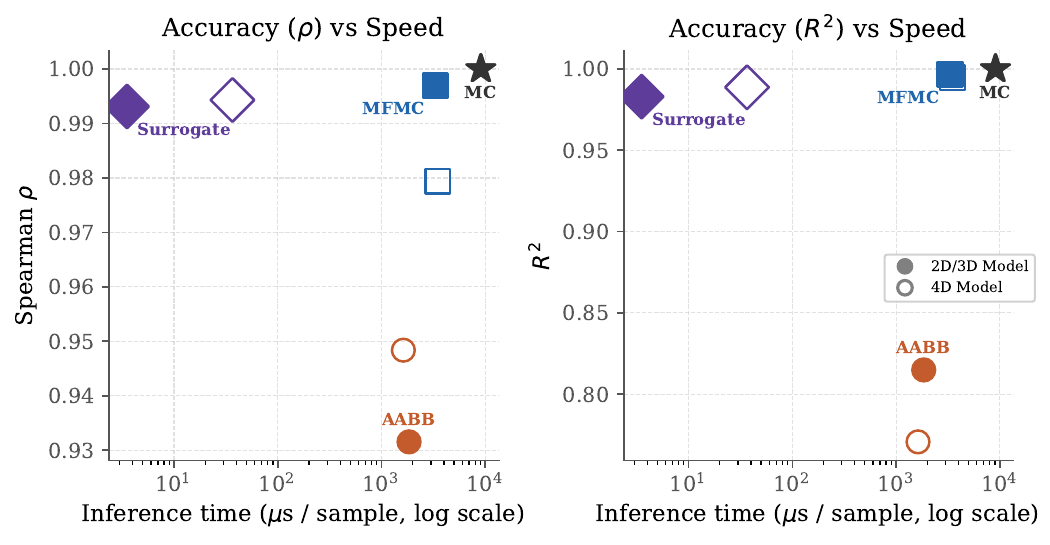}
   \caption{Tradeoff between estimation accuracy and computational cost across all CPS domains. MFMC achieves substantially lower error than AABB while remaining significantly faster than MC.}
    \label{fig:pareto}
\end{figure}

Figure~\ref{fig:pareto} summarizes the resulting accuracy-efficiency tradeoff. 
MC provides the reference estimate at approximately $8998\,\mu s$ per sample.
AABB improves on this at approximately $1846\,\mu s$ but with substantially degraded ranking quality and systematic overestimation. 
MFMC achieves near-reference accuracy at approximately $3306\,\mu s$, trading additional computation for substantially lower estimation error. 
The learned surrogate achieves a $ 935\times$ speedup over MFMC to $3.54\,\mu s$ per sample, while maintaining competitive accuracy. \added[id=TS]{Notably, the MAE of the surrogate is below the standard error of the MC labels.
}

Overall, the results support a hierarchy among estimation paradigms: geometric over-estimators provide an inexpensive but coarse approximation with systematic bias; MFMC achieves near-reference accuracy at higher computational cost; and the learned surrogate uniquely combines competitive accuracy with orders-of-magnitude acceleration, enabling deployment at scales inaccessible to explicit numerical estimators.

\subsection{RQ2: Learning Inconsistency Response Surfaces}
\label{sec:surrogate_model}

While MFMC substantially reduces the computational cost of estimating $I(\theta)$, large-scale exploration of intervention-response spaces still requires repeated numerical evaluations. 
We therefore evaluate whether the learned surrogate model can accurately approximate the inconsistency response landscape to enable scalable exploration across unseen uncertainty configurations.

\begin{table}[t!]
\centering
\caption{Accuracy metrics by zonotope dimension. 95\% bootstrap CIs computed by resampling Saltelli samples.}
\label{tab:acc_by_dim}
\footnotesize
\begin{tabular*}{\columnwidth}{@{\extracolsep{\fill}} l l r r r}
\toprule
Dim & Method & $\rho$ & $R^2$ & MAE \\
\midrule
\multirow{3}{*}{2D} & AABB & 0.932{\tiny$\pm$0.001} & 0.818{\tiny$\pm$0.002} & 0.0514{\tiny$\pm$0.0003} \\
 & MFMC & 0.997{\tiny$\pm$0.000} & 0.997{\tiny$\pm$0.000} & 0.0058{\tiny$\pm$0.0000} \\
 & Surr. & 0.993{\tiny$\pm$0.000} & 0.984{\tiny$\pm$0.000} & 0.0109{\tiny$\pm$0.0001} \\
\midrule
\multirow{3}{*}{3D} & AABB & 0.949{\tiny$\pm$0.003} & 0.780{\tiny$\pm$0.008} & 0.0533{\tiny$\pm$0.0010} \\
 & MFMC & 0.994{\tiny$\pm$0.001} & 0.996{\tiny$\pm$0.000} & 0.0067{\tiny$\pm$0.0002} \\
 & Surr. & 0.989{\tiny$\pm$0.001} & 0.975{\tiny$\pm$0.002} & 0.0150{\tiny$\pm$0.0004} \\
\midrule
\multirow{3}{*}{4D} & AABB & 0.948{\tiny$\pm$0.002} & 0.771{\tiny$\pm$0.009} & 0.0452{\tiny$\pm$0.0010} \\
 & MFMC & 0.979{\tiny$\pm$0.002} & 0.995{\tiny$\pm$0.000} & 0.0068{\tiny$\pm$0.0002} \\
 & Surr. & 0.994{\tiny$\pm$0.000} & 0.989{\tiny$\pm$0.000} & 0.0098{\tiny$\pm$0.0002} \\
\bottomrule
\end{tabular*}
\end{table}

To evaluate generalization behavior, Table~\ref{tab:acc_by_dim} reports estimator performance across different zonotope dimensions. 
The surrogate model maintains consistently strong performance across all evaluated dimensions, with $\rho$ values between $0.989$ and $0.994$ and MAE below $0.015$.
Unlike AABB, whose approximation quality remains largely unchanged across dimensions, the surrogate model maintains near-MFMC performance as dimensions increase.
Notably, the surrogate model achieves its highest rank correlation in the 4D setting, despite having access to only 3 4D training scenarios.
This indicates that the learned representation generalizes effectively beyond the dominant 2D benchmark setting \added[id=JM]{, demonstrating within-domain generalization to unseen scenario configurations.}.

\begin{table}[t]
\centering
\footnotesize
\caption{Inference time per sample and achievable evaluations within fixed time budgets
         for the 2D/3D and 4D surrogate models.}
\label{tab:exploration_budget}
\sisetup{group-separator={,}, group-minimum-digits=4}
\begin{tabular}{l
    S[table-format=4.0] S[table-format=4.0]
    S[table-format=6.0] S[table-format=5.0]
    r r}
\toprule
& \multicolumn{2}{c}{Time ($\mu$s/sample)}
& \multicolumn{2}{c}{Evals in 1\,s}
& \multicolumn{2}{c}{Evals in 60\,s} \\
\cmidrule(lr){2-3}\cmidrule(lr){4-5}\cmidrule(lr){6-7}
{Method} & {2D/3D} & {4D} & {2D/3D} & {4D} & {2D/3D} & {4D} \\
\midrule
MC        & 8998 & 9167 &    111 &   109 &  6.7\,K & 6.5\,K \\
AABB      & 1846 & 1627 &    541 &   614 & 32.5\,K & 36.9\,K \\
MFMC      & 3306 & 3501 &    302 &   285 & 18.1\,K & 17.1\,K \\
Surrogate &    4 &   37 & 283374 & 27329 & 17.0\,M & 1.6\,M \\
\bottomrule
\end{tabular}
\end{table}

Table~\ref{tab:exploration_budget} evaluates the practical exploration capacity enabled by the different estimators.
For the dominant 2D/3D benchmark setting, the surrogate model takes $4\mu$s per evaluation, enabling approximately $2.8\times10^5$ evaluations per second compared to roughly $300$ for MFMC and $540$ for AABB.
Even in the 4D setting, inference remains below $40,\mu$s per sample, allowing more than $27{,}000$ evaluations per second.
Within a one-minute budget, the surrogate enables evaluations up to $17$ million, whereas MFMC remains below $20{,}000$.
These results demonstrate that learned surrogates fundamentally change the feasible scale of intervention-response exploration under realistic computational constraints.

Overall, the results demonstrate that learned surrogates can accurately reproduce the operational consistency boundaries of inconsistency response landscapes. 
At the same time, microsecond-scale inference enables exploration budgets several orders of magnitude larger than explicit MC or MFMC estimation, transforming previously infeasible large-scale intervention analysis into an interactive workflow.

Additionally, an ablation on identical features (full results in the accompanying repository\footnotemark[\value{fnrepo}]) shows the PST improves $R^2$ from $0.974$ for a plain MLP baseline to $0.995$, with the gap widening on the harder 3D holdout ($0.781$ to $0.989$), confirming that the attention and product-set structure contribute beyond the hand-crafted geometric features.

\subsection{RQ3: Sensitivity and Structural Drivers of Inconsistency}
\label{sec:sensitivity_analyis_results}
Beyond predicting inconsistency values, engineers need insights into which uncertainty factors contribute most strongly to inconsistency and how these factors interact. 
We therefore investigate whether the learned surrogate model can accurately reproduce the sensitivity analysis that would otherwise require extensive numerical evaluation.
In particular, we analyze the relative importance of different uncertainty factors, the role of parameter interactions, and how these effects are reflected in the resulting inconsistency response landscapes.

\begin{figure}[t]
    \centering
    \includegraphics[width=0.99\linewidth]{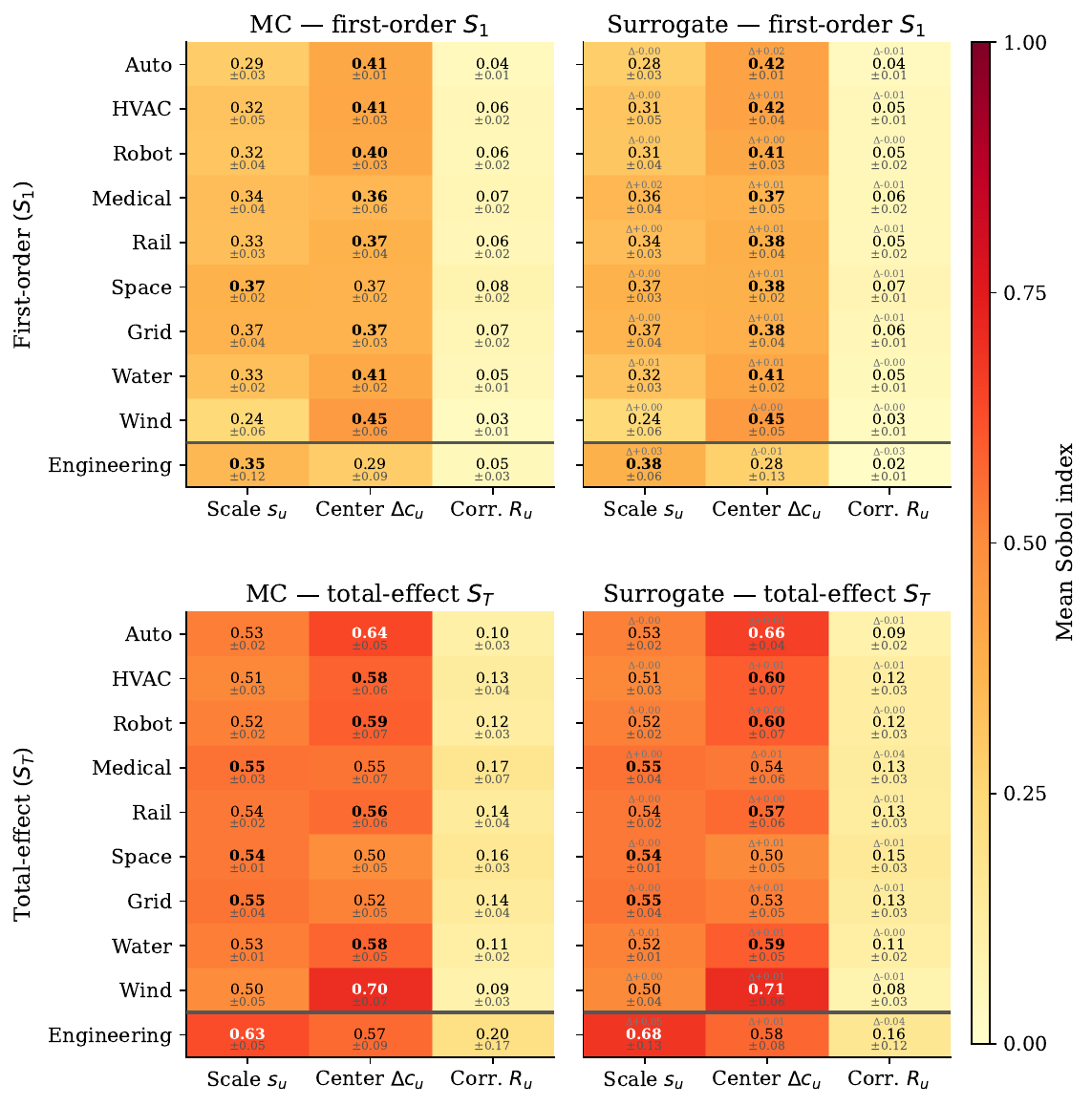}
    \caption{Sobol sensitivity indices for MC and surrogate estimators.}
    \label{fig:sobol_by_domain}
\end{figure}

\begin{figure*}[t!]
    \centering
    \includegraphics[width=0.75\linewidth]{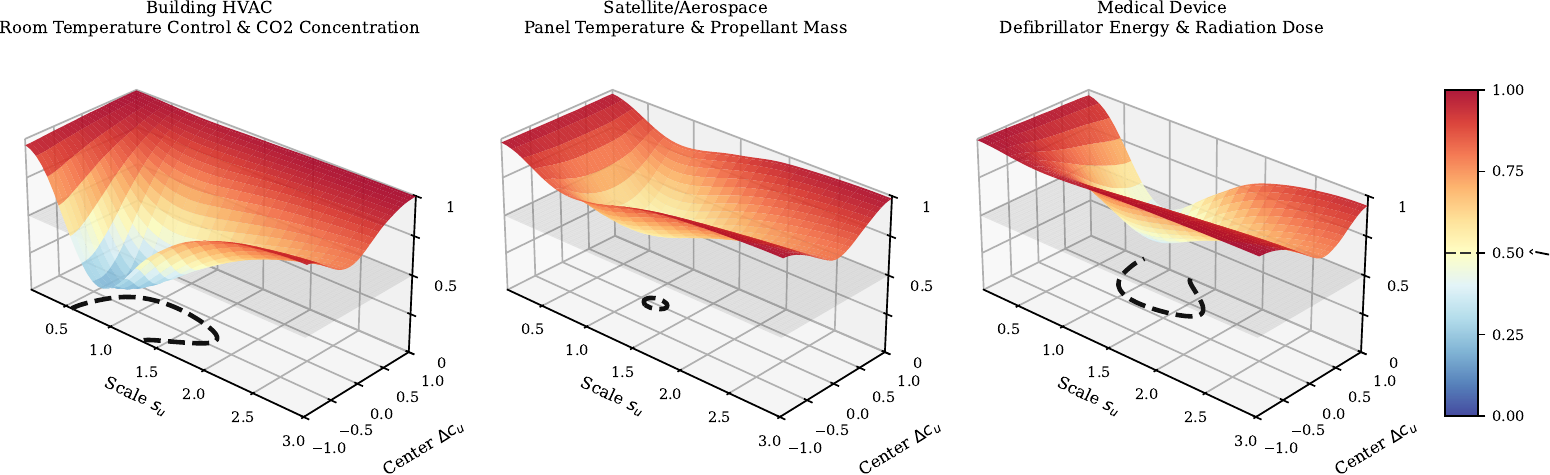}
    \caption{Intervention-response landscapes across representative CPS domains.}
    \label{fig:intervention_landscapes}
\end{figure*}

Figure~\ref{fig:sobol_by_domain} reports first-order ($S_1$) and total-effect ($S_T$) Sobol indices across all evaluated CPS domains. 
Each cell shows the mean index alongside a 95\% bootstrap confidence interval ($\pm$); surrogate panels additionally annotate the signed difference $\Delta$ relative to MC.
The narrow CIs observed across most domains (e.g.\ $\pm 0.01$--$0.02$ for Space and Water) confirm that the reported sensitivity rankings are statistically stable; the wider intervals in the Engineering domain (up to $\pm 0.17$ for $R_u$) reflect the smaller number of available 4D scenarios rather than model instability.

Across nearly all domains, center uncertainty $\Delta c_u$ exhibits the strongest first-order influence, with MC $S_1$ values ranging from $0.36\pm0.06$ (Medical Device) to $0.45\pm0.06$ (Wind Turbine), while scale uncertainty $s_u$ contributes secondary effects between $0.24$--$0.37$. 
Correlation uncertainty $R_u$ remains weak in isolation ($S_1 \leq 0.08$) but exhibits substantially larger total effects ($S_T \leq 0.20$), particularly in the Medical Device and Satellite domains. 
The resulting gap between $S_1$ and $S_T$ indicates that its influence is primarily interaction-driven rather than caused by isolated uncertainty effects.

Importantly, the surrogate reproduces the MC sensitivity structure with very high fidelity across all domains. 
Surrogate--MC differences satisfy $|\Delta| \leq 0.02$ for all first-order indices and $|\Delta| \leq 0.04$ for total effects across the nine CPS domains, with the largest deviation occurring in the Engineering domain ($\Delta S_T(s_u) = +0.05$), where CIs are also widest.
For example, in the HVAC domain, MC yields $(S_1,S_T)=(0.41,0.58)$ for $\Delta c_u$, whereas the surrogate estimates $(0.42,0.60)$. 
Similarly, in the Wind Turbine domain, the surrogate accurately captures the strong center influence ($S_T = 0.71\pm0.06$ vs.\ MC $S_T = 0.70\pm0.07$, $\Delta = +0.01$). 
These results indicate that the surrogate preserves both the global geometry of the inconsistency landscape and its underlying sensitivity structure.

While the Sobol indices identify the dominant uncertainty factors, they do not reveal how inconsistency is distributed across the intervention space.
Figure~\ref{fig:intervention_landscapes}, therefore, visualizes representative intervention-response landscapes.
%

Across all three scenarios, scale and center uncertainty dominate first-order Sobol indices while correlation contributes only marginally (Figure~\ref{fig:sobol_by_domain}); their relative balance differs: Building HVAC is center-led, whereas Medical Device and Satellite/Aerospace show near-equal contributions from both parameters.
Nevertheless, they produce markedly different landscape geometries:
HVAC contains a comparatively broad low-inconsistency region, whereas Medical Device remains highly inconsistent across most of the intervention space and exhibits only a narrow admissible region.
The Satellite scenario represents the most constrained case: the inconsistency surface remains uniformly elevated above~$l$ across the entire intervention space, with no admissible region detectable.

Overall, the results reveal that inconsistency in CPS is strongly interaction-driven and cannot be fully characterized by global sensitivity measures alone.
While Sobol indices identify the dominant uncertainty factors, the intervention-response landscapes show that domains with similar sensitivity patterns can exhibit substantially different consistency regions. 
The learned surrogate preserves both the sensitivity structure and the response-landscape geometry sufficiently well to support scalable sensitivity and intervention analysis.
\subsection{RQ4: Consistency Recourse}
\label{sec:counterfactual-eval}
\begin{figure}[t]
    \centering
    \includegraphics[width=0.68\linewidth]{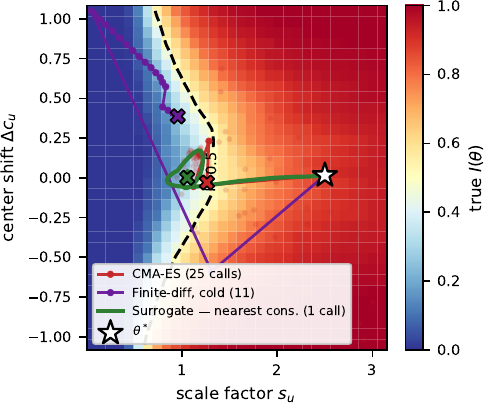}
     \caption{Minimal recourse trajectory for the Engineering domain. The surrogate converges directly to a point near the boundary while the FD makes suboptimal steps and overshoots. CMA-ES reaches a slightly closer point than the surrogate model with significantly more steps.}
    \label{fig:q4_landscape}
\end{figure}
Lastly, we want to answer how well the developed approaches can be used to repair inconsistent CPS by intervening in the uncertainty parameter space. This is important, because depending on the specific CPS, it must be as consistent as possible, or a minimal repair (i.e. small change to $\theta$) to a predefined inconsistency level is required.

We evaluate four different search strategies for that: An evolutionary search (CMA-ES) guided by the MFMC estimator, a finite-difference gradient approximation search with MFMC (FD), a direct gradient-based search with the surrogate model, and a hybrid search where we first search using the surrogate model and then refine up to 10 steps using the FD MFMC estimator.
\added[id=JM]{For the derivative-free baselines (CMA-ES and FD), we set $\lambda = 100$; for the surrogate-based gradient search $\lambda = 42.5$.}All results below are assessed using a full MC estimator (4k samples).

Figure~\ref{fig:q4_landscape} illustrates the search problem. It shows the optimization trajectories of the compared methods (hybrid is not shown since the surrogate alone converged here to a valid point below $I=0.5$). All approaches start from $\theta^*$ and aim to find a configuration with $I<0.5$ that has the smallest possible distance to $\theta^*$. Both the surrogate model and CMA-ES guided by MFMC converge straight to the closest feasible border. However, CMA-ES does this with multiple (25 in total) costly MFMC evaluations. The surrogate slightly overshoots the boundary but then converges close to it inside the feasible region. On the contrary, the FD gradient with MFMC starts into a suboptimal direction and then overshoots widely into the feasible region and must backtrack to the boundary.

\begin{figure}[t!]
    \centering
    \includegraphics[width=0.80\linewidth]{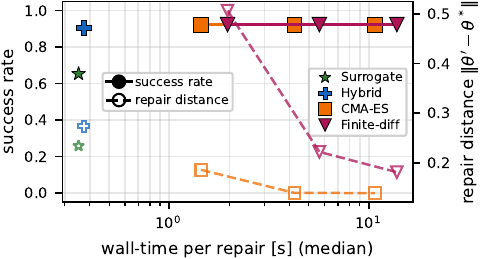}
    \caption{Walltime vs. success rate (filled) and repair distance (hollow) with penalized minimal repair optimization.}   
    \label{fig:minimal_repair_pareto}
\end{figure}

Figure~\ref{fig:minimal_repair_pareto} shows the average success-rate and repair distance against wall-time over 13 held-out evaluation scenarios for the minimal distance repair (i.e., optimizing Eq.~\ref{eq:optim-goal}). Success rate is the number of trials in which the found solution has $I <0.5$, and the repair distance is the distance of the found parameters to the starting parameters. We tested three variants for CMA-ES and FD, 50, 150 or 400 MFMC evaluations as the upper limit. Hollow markers indicate repair distance, and solid ones indicate success rate. The plot shows that all approaches (except the surrogate alone) reach a high success rate of over 90\%. Surrogate and Hybrid converge to a slightly higher repair distance. Yet, this comes at an order-of-magnitude faster runtime. I.e., the surrogate model finds slightly worse repairs, but at roughly 30 ms rather than over 10 seconds.

\begin{figure}[t!]
    \centering
    \includegraphics[width=0.80\linewidth]{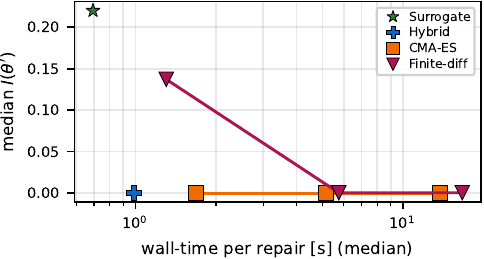}
    \caption{Absolute Runtime vs. median Inconsistency reached for unconstrained repair.}
    \label{fig:min-I-search}
\end{figure}

The results for the unconstrained search for a low $I$ configuration are shown in Figure~\ref{fig:min-I-search}. The x-axis again shows the wall time, and the y-axis shows the median inconsistency reached for all methods, across the same eval scenarios as before. The surrogate alone finds configurations with small $I$,  but significantly higher than all other approaches, which converge to a median of 0, i.e., they find fully consistent configurations. Again, the Hybrid model offers the best tradeoff here, and finds these configurations in almost 
half the runtime of CMA-ES, while FD needs at 150 evals to converge to the same median.

Together, these results demonstrate that both the MFMC and surrogate estimator can be effectively used to search for low-inconsistency configurations, including ones with minimal distance to the initial configuration. While the MFMC-based search offers slightly better results, the surrogate-based gradient search offers a significantly faster runtime.

\section{Related Work}
\label{sec:related-work}
\paragraph{Consistency Analysis under Uncertainty}
Consistency management in multi-model systems is traditionally studied in model-based systems engineering through explicit consistency relations and transformation-based preservation rules~\cite{stevens2017bidirectional, stunkel2021comprehensive}.
These approaches operate on deterministic model states and focus on verifying or restoring consistency for fixed system configurations.
More recent work considers uncertainty-aware consistency analysis~\cite{jongeling2023uncertainty}, but remains centered on local verification rather than scalable analysis of global inconsistency behavior across uncertainty spaces.
%
In contrast, this paper formulates inconsistency as a stochastic response function over uncertainty interventions, enabling scalable estimation, surrogate learning, sensitivity analysis, and intervention-based reasoning over inconsistency landscapes.
\paragraph{Response-Function Approximation} 
Approximating computationally expensive response functions is a central challenge in uncertainty quantification, simulation analytics, and scientific machine learning~\cite{sacks1989design, forrester2008engineering, karniadakis2021physics}. 

Multi-fidelity and variance-reduction methods approximate expensive stochastic estimators by combining inexpensive approximations with high-fidelity evaluations~\cite{peherstorfer2016optimal, peherstorfer2018survey, kennedy2000predicting}. 
Recent work combines surrogate modeling with sensitivity analysis and explainability techniques to enable scalable exploration of complex simulation-driven systems~\cite{saves2025surrogate}. 


Close to our setting, prior work has studied neural prediction of geometric overlap and convex-set properties.
Yuan~\cite{yuan1995neural} proposes one of the earliest neural approaches for measuring the intersection of convex polyhedra, and Bao et al.\cite{bao2023polytopes} demonstrated that geometric polytope properties can be effectively learned from structured representations.
Siamese overlap-prediction networks~\cite{chen2022overlapnet} further provide mechanisms for learning over set-valued geometric inputs. 

In contrast to prior work focusing on pairwise overlap prediction or geometric property estimation, this paper studies inconsistency itself as a stochastic response function induced by uncertainty propagation across interacting multi-model systems.
We combine multi-fidelity approximation with a learned surrogate operating directly on zonotope-based uncertainty representations to estimate inconsistency response surfaces over uncertainty interventions.

\paragraph{Sensitivity Analysis and Feature Importance}
Global sensitivity analysis is widely used to quantify how uncertainty in model parameters affects system behavior. 
Variance-based approaches such as Sobol indices~\cite{sobol2001global} and their estimation via Saltelli sampling~\cite{saltelli2010variance} are commonly applied to identify influential input parameters.
Janzing et al.~\cite{janzing2020feature} formalize feature relevance as a causal problem and argue that meaningful importance measures require interventional rather than purely observational reasoning.
Similarly, Wachter et al.~\cite{wachter2017counterfactual} introduce counterfactual explanations based on minimal interventions that alter model outcomes. Ustun et al.~\cite{DBLP:conf/fat/UstunSL19} frame this as algorithmic recourse: the minimal actionable change that flips an unfavorable outcome.
More recently, Dyer et al.~\cite{dyer2024interventionally} propose interventionally consistent surrogate models that remain valid under distributional interventions.

Building on these perspectives, our work combines global sensitivity analysis with intervention-based reasoning and algorithmic recourse to identify dominant uncertainty drivers of inconsistency and to identify uncertainty interventions that restore consistency.

\section{Conclusion}
\label{sec:conclusion}

We present a scalable framework for estimating, analyzing, and repairing inconsistencies under uncertainty in CPS development.
By reformulating inconsistency as an intervention-response function $I(\theta)$, we enable systematic data generation via MFMC and Saltelli sampling, surrogate-based exploration at the microsecond scale, and consistency recourse.

Our experimental results across 48 scenarios showed that the surrogate model accurately approximates Monte Carlo inconsistency estimates and Sobol indices, models the inconsistency landscape faithfully, while reducing evaluation time from milliseconds to microseconds.
Finally, the proposed recourse approach demonstrates that both MFMC and the surrogate can not only be used to explain inconsistency but also to identify uncertainty modifications that restore consistency. 

Future work will investigate larger multi-model structures and higher-dimensional uncertainty representations.
We also plan to study adaptive sampling strategies that focus on informative regions of the inconsistency response surface, reducing the cost of data generation and surrogate training.

\arxivonly{
\appendices

\section{Surrogate Architecture Ablation}
\label{app:ablation}

\begin{table}[t]
\centering
\footnotesize
\caption{Product-Set Transformer architecture ablation on the 2D/3D split (the 12 held-out 2D/3D scenarios of the 13 held out overall, including the 3D holdout; 200 epochs; metrics at the best validation checkpoint). \texttt{Flat MLP} is the learned baseline on identical features.}
\label{tab:ablation}
\begin{tabular*}{\columnwidth}{@{\extracolsep{\fill}} l r r r r r}
\toprule
Variant & Params & MSE & $R^2$ & $\rho$ & $R^2_{\text{3D}}$ \\
\midrule
PST (full)          & 9{,}394  & 0.00014 & 0.9947 & 0.9963 & 0.9887 \\
Set Transformer     & 10{,}609 & 0.00033 & 0.9875 & 0.9926 & 0.9300 \\
DeepSets            & 8{,}897  & 0.00037 & 0.9859 & 0.9910 & 0.9716 \\
PST, no attention   & 850      & 0.00039 & 0.9849 & 0.9918 & 0.9793 \\
Flat MLP (baseline) & 7{,}937  & 0.00067 & 0.9742 & 0.9857 & 0.7812 \\
\bottomrule
\end{tabular*}
\end{table}

Table~\ref{tab:ablation} ablates the components of the Product-Set Transformer on the 2D/3D split.
The full model (attention plus product head) and the variant without the global head are indistinguishable on aggregate metrics, indicating that the per-dimension factors carry most of the signal.
Removing the product head (\texttt{Set Transformer}) or both the product head and attention (\texttt{DeepSets}) degrades accuracy; removing attention alone (\texttt{PST, no attention}) collapses the model to 850 parameters at a comparable cost, confirming that both the noisy-AND factorization and the attention contribute.
The flat MLP on identical features matches the PST on the 2D scenarios but collapses on the harder, data-scarce 3D holdout ($R^2_{\text{3D}} = 0.781$ vs.\ $0.989$), showing that the architecture's benefit is concentrated where data is scarce and the problem is hardest.
Set Transformer and DeepSets reach their best validation loss within the first two epochs and degrade thereafter; all reported numbers use the best checkpoint.

}
\bibliographystyle{IEEEtran}
\bibliography{bibliography}

@inproceedings{girard2005reachability,
  title={Reachability of uncertain linear systems using zonotopes},
  author={Girard, Antoine},
  booktitle={International workshop on hybrid systems: Computation and control},
  pages={291--305},
  year={2005},
  organization={Springer}
}

@article{schafer2023scalable,
  title={Scalable computation of robust control invariant sets of nonlinear systems},
  author={Sch{\"a}fer, Lukas and Gruber, Felix and Althoff, Matthias},
  journal={IEEE Transactions on Automatic Control},
  volume={69},
  number={2},
  pages={755--770},
  year={2023},
  publisher={IEEE}
}

@inproceedings{althoff2007reachability,
  title={Reachability analysis of linear systems with uncertain parameters and inputs},
  author={Althoff, Matthias and Stursberg, Olaf and Buss, Martin},
  booktitle={46th IEEE Conference on Decision and Control},
  pages={726--732},
  year={2007},
  organization={IEEE}
}

@book{rubinstein2016simulation,
  title={Simulation and the Monte Carlo method},
  author={Rubinstein, Reuven Y and Kroese, Dirk P},
  year={2016},
  publisher={John Wiley \& Sons}
}

@article{peherstorfer2016optimal,
  title={Optimal model management for multifidelity Monte Carlo estimation},
  author={Peherstorfer, Benjamin and Willcox, Karen and Gunzburger, Max},
  journal={SIAM Journal on Scientific Computing},
  volume={38},
  number={5},
  pages={A3163--A3194},
  year={2016},
  publisher={SIAM}
}

@article{saltelli2002making,
  title={Making best use of model evaluations to compute sensitivity indices},
  author={Saltelli, Andrea},
  journal={Computer physics communications},
  volume={145},
  number={2},
  pages={280--297},
  year={2002},
  publisher={Elsevier}
}

@article{scott2016constrained,
  title={Constrained zonotopes: A new tool for set-based estimation and fault detection},
  author={Scott, Joseph K and Raimondo, Davide M and Marseglia, Giuseppe Roberto and Braatz, Richard D},
  journal={Automatica},
  volume={69},
  pages={126--136},
  year={2016},
  publisher={Elsevier}
}

@article{ibarra2023pleiadata,
  title={PLEIAData: consumption, HVAC, temperature, weather and motion sensor data for smart buildings applications},
  author={Ibarra, Antonio Mart{\'\i}nez and Gonz{\'a}lez-Vidal, Aurora and Skarmeta, Antonio},
  journal={Scientific Data},
  volume={10},
  number={1},
  pages={118},
  year={2023},
  publisher={Nature Publishing Group UK London}
}

@inproceedings{goh2016dataset,
  title={A dataset to support research in the design of secure water treatment systems},
  author={Goh, Jonathan and Adepu, Sridhar and Junejo, Khurum Nazir and Mathur, Aditya},
  booktitle={International conference on critical information infrastructures security},
  pages={88--99},
  year={2016},
  organization={Springer}
}

@software{jonkman2024openfast,
  author    = {Jonkman, Bonnie and Platt, Andy and Mudafort, Rafael M. and
               Branlard, Emmanuel and Sprague, Mike and others},
  title     = {{OpenFAST}: v4.0.0},
  year      = {2024},
  publisher = {Zenodo},
  doi       = {10.5281/zenodo.14553563}
}

@book{ericson2004real,
  title={Real-time collision detection},
  author={Ericson, Christer},
  year={2004},
  publisher={Crc Press}
}

@book{hardle2007applied,
  title={Applied multivariate statistical analysis},
  author={H{\"a}rdle, Wolfgang and Simar, L{\'e}opold},
  year={2007},
  publisher={Springer}
}

@article{stunkel2021comprehensive,
  title={Comprehensive systems: a formal foundation for multi-model consistency management},
  author={St{\"u}nkel, Patrick and K{\"o}nig, Harald and Lamo, Yngve and Rutle, Adrian},
  journal={Formal Aspects of Computing},
  volume={33},
  number={6},
  pages={1067--1114},
  year={2021},
  publisher={Springer}
}

@inproceedings{stevens2017bidirectional,
  title={Bidirectional transformations in the large},
  author={Stevens, Perdita},
  booktitle={2017 ACM/IEEE 20th International Conference on Model Driven Engineering Languages and Systems (MODELS)},
  pages={1--11},
  year={2017},
  organization={IEEE}
}

@inproceedings{jongeling2023uncertainty,
  title={Uncertainty-aware consistency checking in industrial settings},
  author={Jongeling, Robbert and Vallecillo, Antonio},
  booktitle={2023 ACM/IEEE 26th International Conference on Model Driven Engineering Languages and Systems (MODELS)},
  pages={73--83},
  year={2023},
  organization={IEEE}
}

@article{peherstorfer2018survey,
  title={Survey of multifidelity methods in uncertainty propagation, inference, and optimization},
  author={Peherstorfer, Benjamin and Willcox, Karen and Gunzburger, Max},
  journal={Siam Review},
  volume={60},
  number={3},
  pages={550--591},
  year={2018},
  publisher={SIAM}
}

@article{kennedy2000predicting,
  title={Predicting the output from a complex computer code when fast approximations are available},
  author={Kennedy, Marc C and O'Hagan, Anthony},
  journal={Biometrika},
  volume={87},
  number={1},
  pages={1--13},
  year={2000},
  publisher={Oxford University Press}
}

@article{sobol2001global,
  title={Global sensitivity indices for nonlinear mathematical models and their Monte Carlo estimates},
  author={Sobol, Ilya M},
  journal={Mathematics and computers in simulation},
  volume={55},
  number={1-3},
  pages={271--280},
  year={2001},
  publisher={Elsevier}
}

@article{saltelli2010variance,
  title={Variance based sensitivity analysis of model output. Design and estimator for the total sensitivity index},
  author={Saltelli, Andrea and Annoni, Paola and Azzini, Ivano and Campolongo, Francesca and Ratto, Marco and Tarantola, Stefano},
  journal={Computer physics communications},
  volume={181},
  number={2},
  pages={259--270},
  year={2010},
  publisher={Elsevier}
}

@inproceedings{janzing2020feature,
  title={Feature relevance quantification in explainable AI: A causal problem},
  author={Janzing, Dominik and Minorics, Lenon and Bl{\"o}baum, Patrick},
  booktitle={International Conference on artificial intelligence and statistics},
  pages={2907--2916},
  year={2020},
  organization={PMLR}
}

@article{sacks1989design,
  title={Design and analysis of computer experiments},
  author={Sacks, Jerome and Welch, William J and Mitchell, Toby J and Wynn, Henry P},
  journal={Statistical science},
  volume={4},
  number={4},
  pages={409--423},
  year={1989},
  publisher={Institute of Mathematical Statistics}
}

@book{forrester2008engineering,
  title={Engineering design via surrogate modelling: a practical guide},
  author={Forrester, Alexander and Sobester, Andras and Keane, Andy},
  year={2008},
  publisher={John Wiley \& Sons}
}

@book{lee2016introduction,
  title={Introduction to embedded systems: A cyber-physical systems approach},
  author={Lee, Edward Ashford and Seshia, Sanjit Arunkumar},
  year={2016},
  publisher={MIT press}
}

@inproceedings{althoff2008reachability,
  title={Reachability analysis of nonlinear systems with uncertain parameters using conservative linearization},
  author={Althoff, Matthias and Stursberg, Olaf and Buss, Martin},
  booktitle={47th IEEE Conference on Decision and Control},
  pages={4042--4048},
  year={2008},
  organization={IEEE}
}

@article{madni2018model,
  title={Model-based systems engineering: Motivation, current status, and research opportunities},
  author={Madni, Azad M and Sievers, Michael},
  journal={Systems Engineering},
  volume={21},
  number={3},
  pages={172--190},
  year={2018},
  publisher={Wiley Online Library}
}

@article{troya2021uncertainty,
  title={Uncertainty representation in software models: a survey},
  author={Troya, Javier and Moreno, Nathalie and Bertoa, Manuel F and Vallecillo, Antonio},
  journal={Software and Systems Modeling},
  volume={20},
  number={4},
  pages={1183--1213},
  year={2021},
  publisher={Springer}
}

@article{makelburg2026surveying,
  title={Surveying uncertainty representation: a unified model for cyber-physical systems},
  author={M{\"a}kelburg, Johannes and Perez-Palacin, Diego and Mirandola, Raffaela and Acosta, Maribel},
  journal={Computing},
  volume={108},
  number={5},
  pages={68},
  year={2026},
  publisher={Springer}
}

@article{nguyen2018cyber,
  title={Cyber-physical specification mismatches},
  author={Nguyen, Luan V and Hoque, Khaza Anuarul and Bak, Stanley and Drager, Steven and Johnson, Taylor T},
  journal={ACM Transactions on Cyber-Physical Systems},
  volume={2},
  number={4},
  pages={1--26},
  year={2018},
  publisher={ACM New York, NY, USA}
}

@article{mordecai2017minding,
  title={Minding the cyber-physical gap: Model-based analysis and mitigation of systemic perception-induced failure},
  author={Mordecai, Yaniv and Dori, Dov},
  journal={Sensors},
  volume={17},
  number={7},
  pages={1644},
  year={2017},
  publisher={MDPI}
}

@article{wachter2017counterfactual,
  title={Counterfactual explanations without opening the black box: Automated decisions and the GDPR},
  author={Wachter, Sandra and Mittelstadt, Brent and Russell, Chris},
  journal={Harv. JL \& Tech.},
  volume={31},
  pages={841},
  year={2017},
  publisher={HeinOnline}
}

@article{karniadakis2021physics,
  title={Physics-informed machine learning},
  author={Karniadakis, George Em and Kevrekidis, Ioannis G and Lu, Lu and Perdikaris, Paris and Wang, Sifan and Yang, Liu},
  journal={Nature Reviews Physics},
  volume={3},
  number={6},
  pages={422--440},
  year={2021},
  publisher={Nature Publishing Group UK London}
}

@article{saves2025surrogate,
  title={Surrogate modeling and explainable artificial intelligence for complex systems: A workflow for automated simulation exploration},
  author={Saves, Paul and Palar, Pramudita Satria and Robani, Muhammad Daffa and Verstaevel, Nicolas and Garouani, Moncef and Aligon, Julien and Gaudou, Benoit and Shimoyama, Koji and Morlier, Joseph},
  journal={arXiv preprint arXiv:2510.16742},
  year={2025}
}

@article{dyer2024interventionally,
  title={Interventionally consistent surrogates for complex simulation models},
  author={Dyer, Joel and Bishop, Nicholas and Felekis, Yorgos and Zennaro, Fabio Massimo and Calinescu, Anisoara and Damoulas, Theodoros and Wooldridge, Michael},
  journal={Advances in Neural Information Processing Systems},
  volume={37},
  pages={21814--21841},
  year={2024}
}

@article{yuan1995neural,
  title={A neural network measuring the intersection of m-dimensional convex polyhedra},
  author={Yuan, Jing},
  journal={Automatica},
  volume={31},
  number={4},
  pages={517--529},
  year={1995},
  publisher={Elsevier}
}

@article{bao2023polytopes,
  title={Polytopes and machine learning},
  author={Bao, Jiakang and He, Yang-Hui and Hirst, Edward and Hofscheier, Johannes and Kasprzyk, Alexander and Majumder, Suvajit},
  journal={International Journal of Data Science in the Mathematical Sciences},
  volume={1},
  number={02},
  pages={181--211},
  year={2023},
  publisher={World Scientific}
}

@inproceedings{froese2026parameterized,
  title={Parameterized hardness of zonotope containment and neural network verification},
  author={Froese, Vincent and Grillo, Moritz and Hertrich, Christoph and Stargalla, Moritz},
  booktitle={International Conference on Learning Representations},
  volume={2026},
  pages={46708--46720},
  year={2026}
}

@article{chen2022overlapnet,
  title={OverlapNet: A siamese network for computing LiDAR scan similarity with applications to loop closing and localization},
  author={Chen, Xieyuanli and L{\"a}be, Thomas and Milioto, Andres and R{\"o}hling, Timo and Behley, Jens and Stachniss, Cyrill},
  journal={Autonomous Robots},
  volume={46},
  number={1},
  pages={61--81},
  year={2022},
  publisher={Springer}
}

@inproceedings{bhave2011view,
  title={View consistency in architectures for cyber-physical systems},
  author={Bhave, Ajinkya and Krogh, Bruce H and Garlan, David and Schmerl, Bradley},
  booktitle={IEEE/ACM second international conference on cyber-physical systems},
  pages={151--160},
}

@inproceedings{DBLP:conf/fat/UstunSL19,
  author       = {Berk Ustun and
                  Alexander Spangher and
                  Yang Liu},
  editor       = {danah boyd and
                  Jamie H. Morgenstern},
  title        = {Actionable Recourse in Linear Classification},
  booktitle    = {Proceedings of the Conference on Fairness, Accountability, and Transparency},
  pages        = {10--19},
  publisher    = {{ACM}},
  year         = {2019},
  doi          = {10.1145/3287560.3287566},
  bibsource    = {dblp computer science bibliography, https://dblp.org}
}

\end{document}